%% file: output.tex
\pdfoutput=1
\documentclass[11pt]{article}

\usepackage[final]{acl}
\usepackage{times}
\usepackage{latexsym}
\usepackage{array}     % for extended tabular functionality
\usepackage{booktabs}
\usepackage{textcomp}
\usepackage[T1]{fontenc}
\usepackage[utf8]{inputenc}
\usepackage{amsfonts}
\usepackage{microtype}

\usepackage{inconsolata}

\usepackage{graphicx}
\usepackage{multirow} 
\usepackage{amsmath}
\usepackage{arydshln}
\usepackage{makecell}
\usepackage{colortbl}
\usepackage{xcolor} % for defining custom colors if needed
\usepackage[normalem]{ulem}
\usepackage{enumitem}
\usepackage{hyperref}

\title{Generation with Dynamic Vocabulary}

\author{
 Yanting Liu\textsuperscript{\rm 1},
 Tao Ji\textsuperscript{\rm 2},  %\\[6pt]
 \textbf{Changzhi Sun\textsuperscript{\rm 1},
 Yuanbin Wu\textsuperscript{\rm 1},
 Xiaoling Wang\textsuperscript{\rm 1}}
\\[6pt]
 \textsuperscript{1} School of Computer Science and Technology, East China Normal University\\
 \textsuperscript{2} School of Computer Science, Fudan University
\\
 \small{
    {%\href{mailto:}{ytliu@stu.ecnu.edu.cn},
    \href{mailto:ytliu@stu.ecnu.edu.cn, ybwu@cs.ecnu.edu.cn, xlwang@cs.ecnu.edu.cn}{\{ytliu@stu,ybwu@cs,xlwang@cs\}.ecnu.edu.cn},
    \href{mailto:taoji.cs@gmail.com, czsun.cs@gmail.com}{\{taoji,czsun\}.cs@gmail.com}
   }
 }
}

\begin{document}
\maketitle

\begin{abstract}
%Vocabulary is a crucial component of language models. Traditional language models generate text by selecting tokens from a fixed vocabulary. In this paper, we introduce a novel dynamic setting for the vocabulary. Under this setting, vocabulary can include arbitrary text spans on demand.
We introduce a new dynamic vocabulary for language models.
It can involve arbitrary text spans during generation.
These text spans act as basic generation bricks, 
akin to tokens in the traditional static vocabularies.  
% Our approach does not necessitate any modifications to language model architecture, as it operates solely on the input and output layers.
We show that, the ability to generate multi-tokens atomically
improve both generation quality and efficiency
(compared to the standard language model, 
%the MAUVE metric increases from 20.47 $\%$ to 25.69$\%$,
the MAUVE metric is increased by $25\%$,
the latency is decreased by $20 \%$).
%Extensive experimental results demonstrate that our approach yields superior generation quality. 
The dynamic vocabulary can be deployed in a plug-and-play way, thus is attractive for various downstream applications.
% Furthermore, we observe that the algorithm of constructing training samples and negative phrases significantly influences the quality of the generated text. 
For example, 
we demonstrate that dynamic vocabulary can be applied 
to different domains in a training-free manner.
It also helps to generate reliable citations in question
answering tasks (substantially enhancing citation results without compromising answer accuracy). 
\footnote{Our source code is publicly available at  
% \url{https://anonymous.4open.science/r/dynamic_vocabulary-7C1C}
\url{https://github.com/Maniyantingliu/generation_with_dynamic_vocabulary}
}
\end{abstract}

\input{intro}
\input{methods}
\input{experiments}

\input{related_work}

\section{Conclusion}
In this paper, we propose a novel approach for dynamically adjusting the model's vocabulary based on the input text. It is a plug-and-play approach that can be simultaneously performed with pre-training tasks. We investigated standard language modeling, domain adaptation, and citation generation, and discussed the impact of different training samples and negative phrase construction methods on the quality of generated text. Our experimental results show that our proposed model can rapidly generate high-quality, high-compression text compared to baselines.

\section*{Limitations}
In this paper, we propose a method to dynamically expand the vocabulary based on the input text. While our approach can improve generation speed and increase the effective length of the generated text, our model does not modify the underlying tokenizer. As a result, it cannot reduce the token numbers for known input information like prompts or questions. The dynamic vocabulary is, therefore, limited to the subsequent content generated by the model.

Furthermore, to obtain embedding representations for phrases, a dynamic phrase encoder is necessary. This encoder has a more intricate structure compared to the model's linear embedding layer and requires additional memory allocation during implementation.

Lastly, our method relies on external techniques, such as a retriever, to obtain relevant documents and extract phrases from them during testing. This adds complexity to the preparation process.

\section*{Acknowledgments}
The authors wish to thank all reviewers for their helpful comments and suggestions. The corresponding authors are Tao Ji, Yuanbin Wu, and Xiaoling Wang. This research was (partially) supported
by National Key R$\&$D Program
of China (2021YFC3340700), NSFC(62076097), the Open Research
Fund of Key Laboratory of Advanced Theory and
Application in Statistics and Data Science (East
China Normal University), Ministry of Education.

\bibliography{custom}
\clearpage
\input{appendix}

\end{document}

%% file: intro.tex
%auto-ignore
\section{Introduction}

% importance of tokenizer
%Vocabulary, while often staying out of the spotlight,
%is a key component of large-scale language models.
%It defines basic bricks (tokens) for composing new sentences,
%bridging different languages \cite{},
%and alleviating toxic usages \cite{}.
%It also influences core model properties
%such as model size and inference cost.
%The generalization (handling OOV tokens)
%and representation (distinguishing different words) ability 
%of vocabulary form the foundation of a successful language model.

%Vocabulary is fundamentally important 
%for large-scale language models.
Vocabulary, which defines basic bricks (tokens) 
for composing new sentences,
bridging different languages,
and alleviating harmful generations,
is essential for language models
\cite{
stahlberg2020neural, lample2019crosslingual, liu2020multilingual,
DBLP:conf/emnlp/KirkBVD22, weidinger2021ethical}.
%To build proper vocabularies, 
%one key problem is 
In modern development, vocabularies are often
obtained by training tokenizers with a pre-defined
vocabulary size on a pre-defined corpus.
%vocabulary size on the specific corpus,
Once built, they are kept unchanged
in the following model construction and deployment
\cite{sennrich2015neural, radford2019language}.

%Conceptually,
%vocabularies containing finer-cut tokens
%can better recognize out-of-domain texts (i.e., smoother generalization),
%while those containing coarse-cut tokens
%can better recognize in-domain words (i.e., more accurate meaning representation).
%To settle the trade-off, most tokenizers determine the
%tokenization granularity by 
%pre-defining a target vocabulary size.
%When the vocabulary is obtained, it is kept fixed
%during the following model construction and deployment.

Though it is sufficient for basic language modeling,
this \emph{static} setting 
makes vocabulary be quietly ignored 
in advanced generation tasks \cite{gao2023enabling, rozière2024code,fried2023incoder,dagan2024getting}.
For example, 
it can not be augmented with new phrases for
better adapting to an unseen domain \cite{koehn2017challenges, jin2020disease, chen2022finqa}
or verbatim reference text spans
for better inline evidence generation \cite{menick2022teaching, gao2023enabling}.
To bring vocabulary back to the stage,
it is natural to ask whether prior constraints
posted by tokenization corpus and fixed vocabulary size 
can be relaxed. 

\begin{figure}
    \centering
    \includegraphics[width=\linewidth]{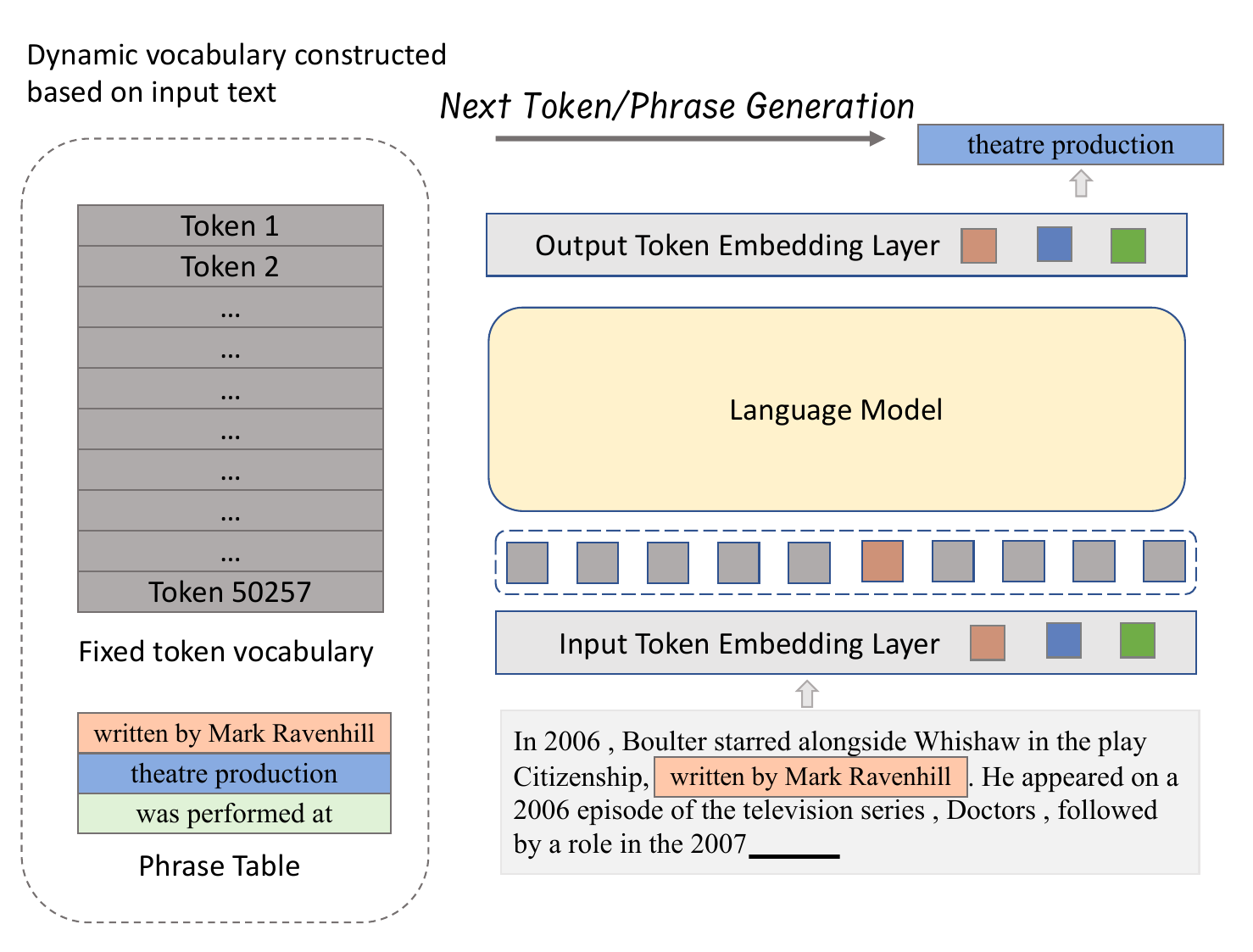}
    \caption{Generation with dynamic vocabulary. 
    The model's vocabulary dynamically changes based on the input text, with phrases serving as basic blocks both for input and output.}
    \label{fig:generation process}
\end{figure}

%Therefore, parallel to other parametric model components,
%it is natural to ask whether vocabulary 

Here, we explore vocabulary in 
a new \emph{dynamic} setting. 
Instead of being a fixed token table, 
dynamic vocabulary is required to be able to include 
\emph{arbitrary text spans} on demand.
%which may include words, phrases, and even sentences
This setup brings new challenges to the language model.
On the input side, 
using a single embedding layer is no longer feasible 
as the full table can not be enumerated.
On the output side, 
the model needs a stronger next-token predictor
as the model allows multiple oracles 
(tokenized to different granularity) for a single string.

In this work, we build a dynamic vocabulary by building 
a dynamic phrase encoder. Akin to the embedding layer,
the encoder maps arbitrary text spans
(called \emph{phrases})
to the input space of language models.
It can be trained with existing language models 
in the same self-supervised manner,
despite that multiple tokens 
(in the original static vocabulary)
can be input or output at a single step.
%We find that, 
Though the paradigm is almost unchanged, 
supporting dynamic tokens
needs non-trivial modification on data curation. 
Specifically, 
we find that, to prevent the learned model from either
biased towards full static token outputs or
towards full new phrase outputs,
it is crucial to make the two properly interleaved 
in training samples.
We also show that the phrase encoder is hard to learn without informative negative samples.
We thus develop two retrieval-based and generation-based 
methods for accelerating the learning of the dynamic phrase encoder.

The obtained dynamic vocabulary
can be deployed in the way of plug-and-play:
the underlying architecture (and backbone parameters) 
of language models 
are kept, and those new on-demand phrases
can be used as ordinary tokens during the generation.
To evaluate the dynamic vocabulary,
we investigate three exemplar applications,
including basic language modeling, domain adaptation,
and generating citations for question answering.
Results show that the new flexibility of vocabulary 
both improve basic generation performances 
(e.g., stronger fluency and diversity scores on WikiText-103 \cite{merity2016pointer}
with lower latency)
and provide a new tool %(orthogonal to existing ones)
to handle advanced language modeling tasks
(e.g., generating more accurate citations with QA scores also increased).

%% file: methods.tex
\section{The Approach}

\subsection{Problem Definition}
\label{section:def}

Given a language model $\mathrm{LM}$, denote $V$ as its vocabulary,
and $x=x_1,x_2,...,x_n$ 
as a tokenized sentence according to $V$
($x_i$ is a token in $V$).
A \emph{dynamic vocabulary} 
 $V'=V\cup P$
augments $V$ with arbitrary phrases (text spans) $P$.
The same sentence
$x$ now can be tokenized to a different
sequence $x_1', x_2',...,x_m'$,
where $x_i' \in V'$.
The usage of dynamic vocabulary $V'$ is 
identical to the vanilla static vocabulary $V$:
the language model $\mathrm{LM}$ 
can accept any token in $V'$ as input and 
choose output tokens from $V'$.

Supporting arbitrary phrase set $P$ and integrating 
$V'$ with language models are two cruxes to implement dynamic 
vocabularies.
For the first one, 
it is possible to support new phrases
by fine-tuning the language model with $V'$,
but it requires updating the model when $P$ changes
which can hardly be used in real applications.
%For the second one,
%it is straightforward to replacing $V$ with $V'$ in training 
%%language model,
We will also see that, for the second crux, 
simply replacing $V$ with $V'$ fails
to learn the language model due to the 
decoding ambiguity introduced by $P$.
We elaborate our solutions in the following sections.

\subsection{Dynamic Phrase Encoder}

Instead of fine-tuning the language model for every possible $P$ to support arbitrary phrase sets,
we build a parametric encoder for those dynamic phrases.
Once the encoder is learned, it can be deployed with the model.

Specifically, 
the dynamic phrase encoder is built with a causal Transformer.
To get the representation of a phrase $p\in P$, 
it first tokenizes $p=w_1,w_2,...,w_s$
according to the static vocabulary $V$,
and after going through several causal Transformer layers followed by an MLP,
the hidden vector of the last token $\mathbf{h}_s$
is the vector representation of $p$.

The above setting is different from existing work in three ways \cite{lan2023copy,teehan2024college}.
First, it is common to use a Transformer encoder
(full attention) to build the phrase encoder,
while we apply a Transformer decoder (causal masking).
The choice is mainly guided by efficient negative sampling 
(see Section \ref{section:training} for further details).

Second, the dynamic phrase encoder adopts 
the same tokenizer of $\mathrm{LM}$ (which is used to build the 
static vocabulary $V$).
Sharing tokenizers means the language model doesn't need to
load additional vocabularies and tokenizers during inference.
\footnote{As a comparison, 
the phrase encoder in CoG \cite{lan2023copy} is BERT, and one should load both the BERT vocabulary and GPT-2 vocabulary when testing.}

Third, to further unify the new phrase encoder and the LM,
%following the same idea of consistent treatment, 
we use a non-contextualized representation of phrases,
which makes the new phrases more like the original tokens in $V$.
Contextualized representations can also be 
used \cite{joshi-etal-2020-spanbert,lan2023copy},
but it means that, besides the phrases themselves,
the contexts of them should also be included
in the dynamic vocabulary.

To summarize, the considerations above aim to make
the dynamic phrase encoder align with the 
embedding layer as much as possible:
both of them map tokens (phrases) into the input space
of the language model,
one by lookup operations, and another by running the phrase encoder.

%\paragraph{Phrase Encoder.} We employ a causal attention Transformer to compute the vector representations of the phrases. The context-independent phrase representations are obtained as follows. For a set of phrases ${P^1, P^2, ..., P^n}$, we apply the transformer to obtain last token hidden state representations $H^{n\times d}$, followed by projecting $H^{n \times d}$ to the same dimension, as phrase embeddings. The advantages of the above representation method are that (1) we maintain the same content-independent phrase representation as the single token representations; and (2) It is convenient to incorporate negative examples.

\subsection{Inference with Dynamic Vocabulary}

In testing time, the new dynamic vocabulary can be used
as the ordinary vocabulary.
We take an auto-regressive language model $\mathrm{LM}$ as an example.
For a set of new phrases $P$
\footnote{The phrase set $P$ can change at each decoding step.
Here, for simplicity, we assume it is kept unchanged during 
testing, and we can run the dynamic phrase encoder only once.},
we run the learned dynamic phrase encoder to get 
representations of its phrases, denoted by a matrix
$\mathbf{P}$. 
The language model's input and output embedding matrices 
$\mathbf{W}_\mathrm{emb, in}, \mathbf{W}_\mathrm{emb, out}$
are expanded with these embeddings,
\begin{align*}
     \mathbf{W}'_\mathrm{emb, in} &= [\mathbf{W}_\mathrm{emb, in}, \mathbf{P}], \\ 
     \mathbf{W}'_\mathrm{emb, out} &= [\mathbf{W}_\mathrm{emb, out}, \mathbf{P}].
\end{align*}

At each auto-regressive decoding step, 
the language model $\mathrm{LM}$ outputs a hidden
vector $\mathbf{h}_{<i}$ representing current prefix
$x'_{<i}$, the probability of next token is
\begin{align}
\label{eq:log-prob}
    & \mathbb{P}(x'_i=k|x'_{<i}) = 
    Z^{-1}\exp(\mathbf{h}_{<i}\cdot \mathbf{e}_\mathrm{out}^k) \\ 
    & Z = \sum_{k'\in V}
    \exp(\mathbf{h}_{<i}\cdot\mathbf{e}_\mathrm{out}^{k'}) +
    \sum_{k'\in P}
    \exp(\mathbf{h}_{<i}\cdot\mathbf{e}_\mathrm{out}^{k'}), \nonumber
\end{align}
where $\mathbf{e}_\mathrm{out}^k$ is the $k$-th column 
of $\mathbf{W}'_\mathrm{emb, out}$.
When the $i$-th token is selected, no matter whether it is a token
in $V$ or a phrase in $P$, its embedding is looked up from
$\mathbf{W}'_\mathrm{emb, in}$ 
as the input of the next decoding step.
\footnote{When decoding a phrase, another option adopted by \cite{joshi-etal-2020-spanbert,lan2023copy}
is to unfold tokens in the phrase and input
them individually. Despite the inconsistency between input and output vocabulary (our experiments indicate a negative influence on performances), this setting may also 
slow the decoding speed (or generate shorter texts given a fixed length budget)
even if it can predict a phrase.}

\begin{figure*}
    \centering
    \includegraphics[width=\linewidth]{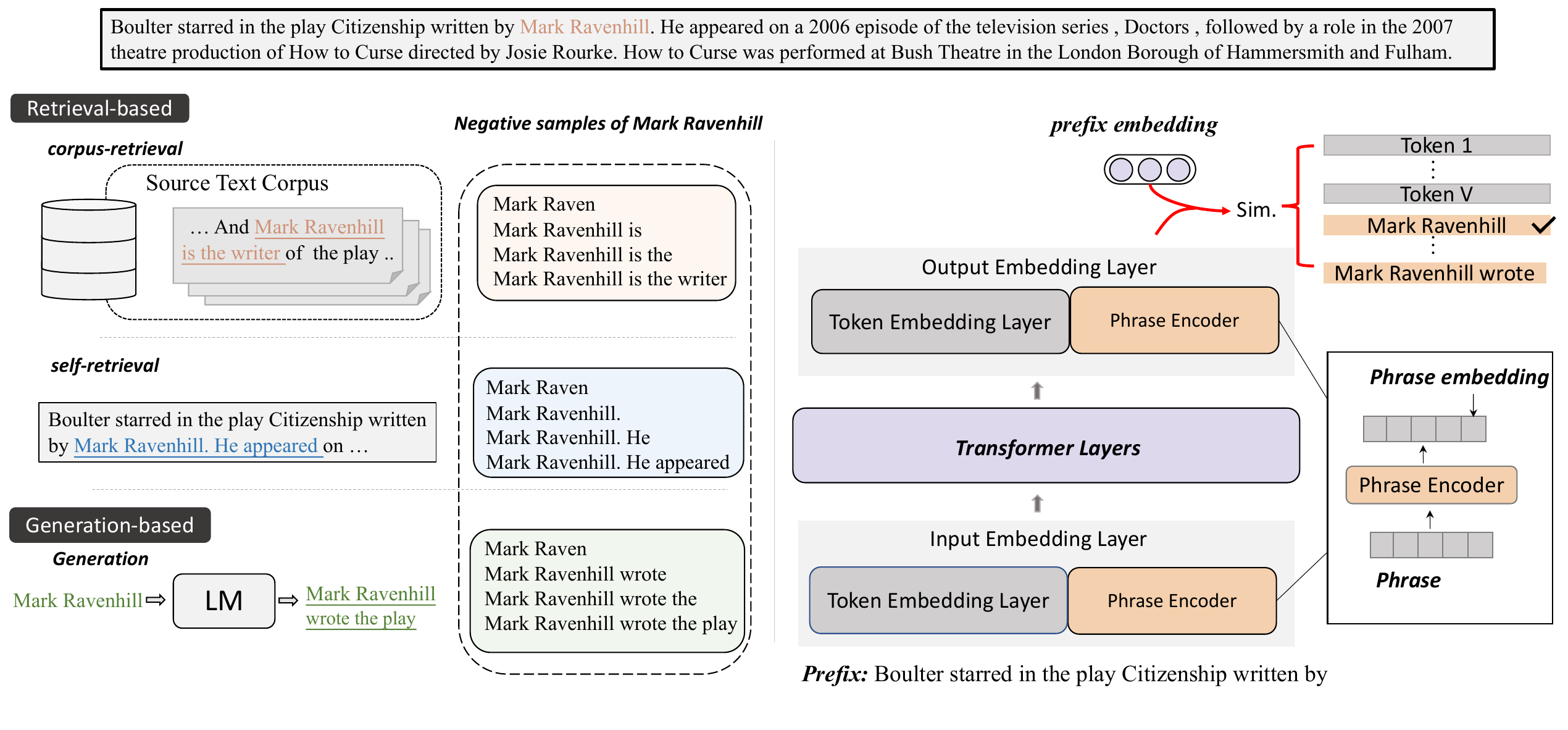}
    \caption{The overall architecture of our proposed dynamic vocabulary. During training, there are four sources of negative phrases: pre-batch, corpus-retrieval, self-retrieval, and generation. Phrases are embedded by the dynamic phrase encoder with an additional linear layer. The hidden layer of the last token serves as the phrase embedding. In the model input layer, phrases are treated as a basic brick without splitting into tokens.}
    \label{fig:overall_architecutre}
\end{figure*}

%Our proposed model mainly consists of two components: a context-independent phrase encoder and a prefix encoder.

\subsection{Training with Dynamic Vocabulary}
\label{section:training}

\paragraph{Building Samples}
To train the dynamic phrase encoder, 
we follow the same self-supervision regime as the training
of language models.
The key difference here is that, besides tokens in $V$,
we need to organize phrases (text spans) in a training sample
for learning the phrase encoder.
In particular,
1) the diversity of training-time in-domain phrases would 
influence the generalization of the learned phrase encoder,
and 2) the distribution of phrases in samples would influence how the language model switches between tokens and phrases. 

For building phrases, 
we test the following two methods. %FMM, N-words, and N-ids.
\begin{itemize}[leftmargin=*]
  \item \emph{``real'' phrases}.
  We can use classical chunking algorithms to recognize phrases in 
  a sentence. The resulting phrases can be recognized as single grammatical units or as common word collocations.
  Here, we follow \citet{lan2023copy} to use an unsupervised
  chunker forward maximum matching (FMM).
  Basically, FMM recognizes phrases that frequently appear
  in a support corpus and as long as possible. 
  The algorithm (and other external chunkers)
  may need additional time costs to compile samples
  (e.g., in our experiments, FMM needs $\approx 15$ hours to 
  build its phrase table).
  \item \emph{Ngrams}.
  Another candidate set of phrases is ngrams,
  which is much simpler to build than involving 
  external chunkers. 
  Though a ngram may not carry a meaning,
  it could be a stronger learning target for the phrase encoder:
  the connections between ngrams and its contexts are more complex 
  than ``real'' phrases (as they usually follow the simple patterns
  which are used to extract them).
  We study two settings, ngrams of words and ngrams of tokens
  (denoted by N-words and N-ids respectively).
  Taking N-words as an example, a word tokenizer 
  \footnote{N-words uses the word tokenizer in the NLTK toolkit, 
  and N-ids uses GPT-2's tokenizer.}
  first recognizes words in a sentence,
  then randomly sequences of $2$-$5$ consecutive words 
  are grouped into phrases. 
\end{itemize}

Next, given a sentence and a set of candidate phrases, 
we need to determine the distribution of phrases.
One may build samples with full ngrams phrases,
but they could be both hard to learn (the learning ignores
the prior knowledge of original vocabulary $V$ in the model), 
and hard to apply (the setting is rare in applications).
In our practice, 
to accelerate learning and prevent unnecessary data bias,
it is crucial to make phrases and tokens properly interleaved in training samples.
Therefore, we control the interval between 
two phrases to be at least five tokens.

\paragraph{Negative Phrases}
After building training samples, we can directly 
optimize the log-probability defined in Equation \ref{eq:log-prob},
which requires the correct next token in $V'=V\cup P$
has the largest logit than 
other tokens in $V$ and $P$ (negative tokens).
However, the number of phrases in the training set 
would be large, and it is prohibitive to include all of them
in the loss function.
\footnote{It is worth noting that all training time phrases
are dropped after learning the encoder.
For ngram phrases (N-words and N-ids), 
phrases are built on the fly in the batching process,
and there is no global training time $P$.}
A common workaround is to include only in-batch
and pre-batch phrases in $P$ \cite{gao-etal-2021-simcse}.
Unfortunately, it doesn't help learning the phrase encoder.
Specifically, we find that the model struggles
to correctly transit from a phrase token to 
an ordinary token and vice versa.
More concretely, when predicting a phrase $p=w_1, w_2,...,w_s$,
the dynamic phrase encoder has trouble on
distinguish $p$ from
1) phrases which are prefixes of that phrase 
(e.g., $w_1w_2$ and $w_1w_2w_3$)
and 2) phrases which have $p$ as their prefix 
(e.g., $pw_{s+1}$ and $pw_{s+1}w_{s+2}$).
Therefore, we also manually add the above phrases
to $P$ in each batch (we call them informative negative phrases).

For the first type, we can simply enumerate all prefixes of $p$.
For the second type, we develop \emph{retrieval-based} and 
\emph{generation-based} methods 
for getting successor tokens of $p$,
\begin{itemize}[leftmargin=*]
    \item retrieval-based continuation finds appearances
    of $p$ in a support corpus and takes $p$ and 
    its successor tokens there as negative phrases 
    (\textbf{corpus-retrieval}).
    \footnote{Due to the time complexity of matching phrases, 
    we only adopt corpus-retrieval when phrases are obtained by FMM,
    and keep the efficiency of Ngram phrases. }
    One simplification is only considering $p$'s successor tokens 
    in the current sample (\textbf{self-retrieval}).
    \item generation-based continuation, instead of searching corpus,
    tries to get synthetic negative phrases by employing 
    a language model.
    \footnote{Here we use GPT-2, stronger models can also be applied.}
    The model is prompted with $p$ and the following generations are included in $P$ 
    (\textbf{generation}).
\end{itemize}

Finally, regarding getting embeddings of these
informative negative phrases, recall that we adopt an 
causal Transformer as the phrase encoder and use the hidden state
of the final token to represent $p$,
the embeddings of negative phrases could be efficiently obtained
by feeding the longest phrase to the encoder.

\paragraph{Loss Functions}

The first part of the training loss is defined
by Equation \ref{eq:log-prob} (with negative samples
added to $P$), 
which we denote by $L_p$. 
We also add a special setting of $L_p$ in the loss
(denoted by $L_t$),
in which $P=\emptyset$ (i.e., the vanilla language modeling).
It helps to maintain generation ability with 
the static vocabulary $V$.

We can further align the above two settings
by requiring their next token distributions at each 
token position are close (measured by KL divergence).
Concretely, given a sentence $x$,
recall that (Section \ref{section:def})
the oracle of training $L_p$ is $x'_1, x'_2,..., x'_m$, 
the oracle of training $L_t$ is $x_1, x_2,..., x_n$.
Assume a function $\sigma$ which aligns
$x'_i$ to a token position in $L_t$'s oracle: 
if $x'_i$ is a token in $V$,
it is mapped to the same token position, 
otherwise, $x'_i$ is mapped to its last token's position.
\begin{equation*}
    L_{kl} = \frac{1}{m} \sum_{i=0}^m\mathrm{KL}
    (\mathbb{P}(x'_i|x'_{<i})||
    \mathbb{P}(x_{\sigma(x'_i)}|x_{<\sigma(x'_i)})).
\end{equation*}
The final loss function is $L = L_p + L_t + L_{kl}$.

%% file: experiments.tex
\section{Experiments }

\subsection{Setups}

\paragraph{Configurations}
% In this study, we conducted training on our model using the Wikitext-103 dataset and employed GPT-2 \cite{radford2019language} for the initialization to separately construct prefix encoder and phrase encoder. The training was carried out on two NVIDIA RTX 3090 GPUs, each with 24GB of memory, over a total of 400,000 training steps. During the training process, we implemented a gradient accumulation step of 2, with a batch size of 4. We also used a linear learning rate schedule with a warmup, alongside the AdamW optimizer \cite{loshchilov2019decoupled}, maintaining the default beta values. The initial learning rate was set at 5e-5. Additionally, we applied gradient clipping with a clipping value of 1.0 to ensure training stability.
% For the testing phase, in order to validate the effectiveness of our method, we need to construct appropriate phrases for each test sample. To do this, we utilized the DPR \cite{DBLP:conf/emnlp/KarpukhinOMLWEC20} model to calculate semantic vectors and employed the FAISS technique to retrieve the top k most similar documents from the training set based on the current sample. The value of k is 32. Subsequently, we considers all n-grams with them as candidate phrases. It is worth mentioning that our proposed method is applicable to both greedy sampling and nucleus sampling strategies. When conducting nucleus sampling, we set the p value to 0.95.

 For a fair comparison with baselines, we use GPT-2 \cite{radford2019language} to initialize both the language model and dynamic phrase encoder. To collect phrases for each test sample, $k$ related documents are retrieved by the semantic matching model, DPR \cite{DBLP:conf/emnlp/KarpukhinOMLWEC20} and the vector search toolkit, FAISS \cite{johnson2019billion}. In our paper, the value $k$ is set to 32. 
 
 We experiment with several negative sampling and sample-building methods and set N-words with ``self-retrieval + generation'' as default. Besides, we initialize the language model with two models of different scales, GPT-2 and Tinyllama \cite{zhang2024tinyllama}, to verify the effectiveness of our proposed method. We employ full-parameter fine tuning for GPT-2 and LoRA \cite{hu2021lora} for Tinyllama. 
% In testing time, %to validate the effectiveness of our method,
% we construct phrases for each test sample. 
% By default, we follow CoG \cite{lan2023copy} to use DPR
% to retrieve the top-k ($k=32$)
% relevant documents from the training set.
% We consider all n-grams of top-k documents as candidate phrases. 
% It is worth mentioning that our proposed method is applicable to both greedy sampling and nucleus sampling strategies. When conducting nucleus sampling, we set the $p$ to 0.95.
Please refer to Appendix \ref{ref:IMPLEMENTATION DETAILS of baselines} for more details.

\paragraph{Baselines} 
We compare the proposed method with the following state-of-the-art models as baselines:

{\bf Transformer} \cite{vaswani2023attention} is the standard token-level language model. We fine-tune the pre-trained GPT2 in our experiments.

{\bf KNN-LMs} \cite{khandelwal2020generalization} extends a pre-trained neural language model by linearly interpolating it with a k-nearest neighbors(KNN) model.

{\bf RETRO} \cite{borgeaud2022improving} is a retrieval-enhanced transformer that combines a frozen Bert retriever, a differentiable encoder, and a chunked cross-attention mechanism.

{\bf CoG} \cite{lan2023copy} decomposes text generation into a series of copy-and-paste operations. It first retrieves semantically relevant documents and then considers all n-grams within them as candidate phrases
\footnote{CoG adopts a two-stage search strategy (document retrieval followed by phrase extraction) while CoG-2 \cite{cao2024retrieval} generates text directly through phrase retrieval. However, CoG-2 fails to provide any code, thus precluding any comparative analysis.}.

{\bf MWT} \cite{Gee_2023} propose to expand vocabulary with top-k frequent n-grams in support corpus. 
Rather than expanding vocabulary dynamically, 
it still focuses on building a static vocabulary.

%{\bf MWT} \cite{Gee_2023} proposed a sequence compression approach, which reduces textual inputs by expanding vocabulary with top-k frequent n-grams in the given corpus. 
% exploiting the use of multi-word expressions drawn from the training set according to their top-K frequencies.

\paragraph{Metrics} 
 We use four automatic evaluation metrics to measure the quality of the generated texts \cite{lan2023copy, cao2024retrieval},: (i) {\bf MAUVE} \cite{pillutla2021mauve} measures the distribution similarity between the reference text and generated text; (ii) {\bf Rep-n} \cite{welleck2019neural} reflects the repetition at different n-gram levels in the generated text; (iii) {\bf Diversity} \cite{welleck2019neural} evaluates the variety of generated content; and (iv) {\bf Perplexity} measure the difficulty in predicting the next word in a sequence. In addition, we also compare the average time cost of different methods to decode a continuation consisting of 128 tokens given a prefix of 32 tokens, referred to as {\bf latency}. The details for these metrics can be found in Appendix \ref{appendix: automatic evaluation}

% \begin{itemize}
%     \item {\bf MAUVE} \citet{pillutla2021mauve} measures how closely the token distribution in generated text matches that in human-writted text across the entire test set. We follow prior work and leverages the GPT2-large model to generate the scores. In our implementation, the scaling factor is set as 2.0.
%     \item {\bf Rep-n} \citet{welleck2019neural}  measures the repetition at different n-gram levels in generated text. It is defined as $100 \times (1.0 - \frac{|unique n-gram(x)|}{|total n-gram(x)|})$. Higher the Rep-n represents the severe degeneration problem in generations.
%     \item {\bf Diversity} \citet{welleck2019neural} evaluates the variety of generated content, which is formulated as $\prod_{n=2}^4(1- \frac{Rep-n}{100})$.More informative generations gets higher Diversity scores.
%     \item {\bf Perplexity} is a measure of the uncertainty or difficulty in predicting the next word in a sequence. A lower perplexity score indicates that the model is more certain about its predictions.
% \end{itemize}

%\subsection{Results}

We investigate three applications: basic language modeling, domain adaptation, and generating citations for question answering.

\subsection{Basic Language Modeling} \label{wikitext103-results}
% WikiText-103 \cite{merity2016pointer}, a large-scale dataset comprising over 100 million words extracted from a diverse set of Wikipedia articles, has become a standard benchmark for assessing the performance of language models. 
% WikiText-103 \cite{merity2016pointer}, comprising over 100 million words from Wikipedia articles, is a standard benchmark for assessing the performance of language models. 

\begin{table*}[htp]
\begin{center}
\scalebox{0.85}{
\begin{tabular}{cccccccc}
    \toprule 
\multicolumn{1}{c}{\bf Model}  & \multicolumn{1}{c}{\bf MAUVE $\uparrow$}  & \multicolumn{1}{c}{\bf Rep-2 $\downarrow$} & \multicolumn{1}{c}{\bf Rep-3 $\downarrow$} & \multicolumn{1}{c}{\bf Rep-4 $\downarrow$} & \multicolumn{1}{c}{\bf Diversity $\uparrow$} & \multicolumn{1}{c}{\bf Latency(s)$\downarrow$} & \multicolumn{1}{c}{\bf PPL $\downarrow$}\\
\midrule
\bf Transformer & 20.47 & 41.96 & 36.82 & 33.74 & 24.30 & 1.10 & \bf 3.60 \\
\bf RETRO & 19.59 & 43.78 & 38.58 & 35.35 & 22.33 & 4.43 & 3.96 \\
\bf KMM-LM$^*$ &  19.92 & 43.79 & 38.76 & 35.69 & 22.13 & 10.36 & 3.48 \\
\bf CoG & 21.61 & 34.77 & 30.67 & 28.35 & 32.41 & 1.04 & 7.89 \\ 
\bf MWT & 24.74 & 33.78 & 26.72 & 22.76 & 37.48 & 1.13 & 5.58 \\ 
\bf Ours & \bf 25.69 & \bf 27.77 & \bf 20.80 & \bf 17.08 & \bf 47.44 & \bf 0.99 & 8.03 \\
\bottomrule
\end{tabular}}
\end{center}
\caption{The automatic evaluation on the test set of WikiText-103. $*$ indicates that we directly utilize the results from the CoG paper for KNN-LM due to limited GPU memory. Additionally, our method retrieves only 32 documents for phrase segments during evaluation, whereas CoG retrieves 1024. \citet{Gee_2023} apply MWT to encoder-only model but we implement MWT with GPT-2.}
\label{table: automatci evaluation on wikitext103}
\end{table*}

We use GPT-2 and WikiText-103 \cite{merity2016pointer} for
evaluating open-ended language generation.
For each test sample, we provide the first $32$ tokens 
as a context prefix, 
and both the baselines and our model will generate the subsequent 128 tokens 
(tokens are in GPT-2's original vocabulary). 
%For a fair comparison, our model generates texts of the same length as other baselines 
%(the actual token number is larger than $128$ since each decoding step can output multiple tokens).

% This evaluation methodology allows for a fair comparison of the models' ability to generate coherent and contextually relevant text in an open-ended setting. 
%\paragraph{Results.} 
% Table \ref{table: automatci evaluation on wikitext103} showcases the results of our method in comparison to all baselines on the WikiText103 corpus, with a focus on automatic evaluation.
The results are listed in Table 
\ref{table: automatci evaluation on wikitext103}. 
We find that,
\begin{itemize}[leftmargin=*]
    \item Regarding generation quality, 
    language models with dynamic vocabulary 
    can outperform standard Transformer with $5.22\%$
    MAUVE score (better fluency).
    %While MWT gets $24.74\%$ MAUVE, it is a static vocabulary. 
    Meanwhile, our model achieves $47.44\%$ diversity, 
    which is much better than other baselines. 
    \item Regarding generation efficiency,
    dynamic vocabulary achieves the best latency. 
    %Given that phrase representations are pre-computed, encoding times are excluded. 
    %As seen, our model exhibits a clear advantage in faster text generation, nearly 0.99s to generate 128 ids.
    The reason is that a single phrase contains several tokens, which translates to fewer decoding steps for a given decoding length budget. 
    % Compared with CoG, while CoG also outputs phrases,
    % it still runs at a token level, which leads to a higher
    % latency than our model.
    
    \item the perplexity of dynamic vocabulary (our model
    and CoG) is higher than that of the Transformer. 
    This discrepancy could potentially stem from the fact that during testing, the input prefixes are strictly composed of tokens from a fixed vocabulary, whereas the model is not subjected to such constraints during training, which results in an inconsistency between the training and testing data distributions, potentially leading to the observed difference in perplexity scores.
\end{itemize} 
 %We also evaluate the generation results under nucleus sampling (p = 0.95), with detailed metrics provided in the appendix \ref{appendix: nucleus results}. 
 
 % We also evaluate the generation results under nucleus sampling in Appendix \ref{appendix: nucleus results}, conduct experiments on TinyLlama in Appendix \ref{appendix: scale to 1B}, and attempt real-time adaptability in Appendix \ref{appendix: Real-time adaptability}. 
  We also evaluate the generation results under nucleus sampling and attempt real-time adaptability. 
  The details are located in Appendix \ref{appendix: nucleus results}, \ref{appendix: Real-time adaptability} separately. Moreover, the analysis of memory and computational resources occupation during inference can be found in Appendix \ref{appendix: Memory and computational resources}.

\paragraph{Human Evaluation} 
% To ensure the reliability of the aforementioned automated evaluation and the quality of the generated text,

% \begin{table}
% \begin{center}
% \scalebox{0.8}{
% \begin{tabular}{lccc}
% \hline
%  \parbox{3cm}{\multicolumn{1}{l}{\bf Ours versus  ($*$)}} & \multicolumn{1}{c}{\bf Better} & \multicolumn{1}{c}{\bf No Prefer}  & \multicolumn{1}{c}{\bf Worse} \\
% \hline
% \bf{Human Evaluation} & & & \\
% Transformer & 0.57 & 0.22 & 0.21\\
% MWT & 0.55 & 0.21 & 0.24\\
% CoG & 0.53 & 0.22 & 0.25 \\
% \hline
% \bf{GPT-4 Evaluation} & & & \\
% Transformer & 0.61 & 0.05 & 0.34\\
% MWT & 0.58 & 0.02 & 0.40 \\
% CoG & 0.58 & 0.08 & 0.34 \\
% \hline 
% \end{tabular}}
% \end{center}

\begin{table}
% \scalebox{0.8}{
\centering
\small
\begin{tabular}{lccc}
\toprule
 {\bf Ours versus  ($*$)} & \bf Better & \bf No Prefer  & \bf Worse \\
\midrule
 \multicolumn{4}{c}{\bf{Overall Evaluation}}  \\
        
Transformer & 0.57 & 0.22 & 0.21\\
MWT & 0.55 & 0.21 & 0.24\\
CoG & 0.53 & 0.22 & 0.25 \\
\midrule
 \multicolumn{4}{c}{\bf{Comparasion in * aspect}} \\
Fluency & 0.41 & 0.31 & 0.28\\
Coherence & 0.44 & 0.28 & 0.28 \\
Informativeness & 0.56 & 0.18 & 0.26 \\
Grammar & 0.32 & 0.43 & 0.25\\
\bottomrule
\end{tabular}
% }
\caption{Overall human evaluation on WikiText-103 and detailed comparison with GPT-2 in the four aspects. In the overall evaluation, we regard the four aspects as a whole and hence there is a single score. ``Better'' represents that our proposed model's output is superior; ``No prefer'' indicates that the performance is comparable; and  ``worse'' denotes that our model's output is inferior.}
\label{gpt evaluation on wikitext103}
\end{table}

To gain further assessment,
we also run human evaluation on a random sample of 100 generations. 
% The process is similar to the CoG paper's evaluation pipeline: 
For each test sample prefix, the annotators are given two continuations generated by the baseline and our model respectively in random order. Annotators are asked to choose which one is better (in terms of fluency, coherence, informativeness, and grammar). When annotators make different decisions on the same sample, they will discuss and make the final decision. We regard the four aspects as a whole in the overall evaluation and also score in each aspect. As shown in Table \ref{gpt evaluation on wikitext103}, dynamic vocabulary outperforms the Transformer with better cases of 57 and 21 cases of slight inferiority and wins more cases in all four aspects, especially coherence and informativeness.
The results are consistent with MAUVE, which shows that the model with dynamic vocabulary possesses a stronger generation capability and the outputs from our method often have a tighter connection with the preceding text.

%Although human evaluation is considered the gold standard for assessing human preferences, it is slow and costly. \citet{zheng2023judging} have demonstrated that strong LLMs, such as GPT-4, can match most human preferences well. Therefore, 
We also employ GPT-4 \cite{openai2024gpt4} 
%(GPT-4-0125-preview) 
for further assessment. Detailed implementations and prompts are in Appendix \ref{GPT2 evaluation}. 
The results are consistent with the aforementioned evaluations.

\paragraph{Case Study} 

To provide more proof of the effectiveness of our proposed model and the quality of its generation, we conduct some case studies and compare texts generated by our proposed model and GPT-2. 
\begin{figure*}[htbp]
\centering
\includegraphics[scale=0.60]{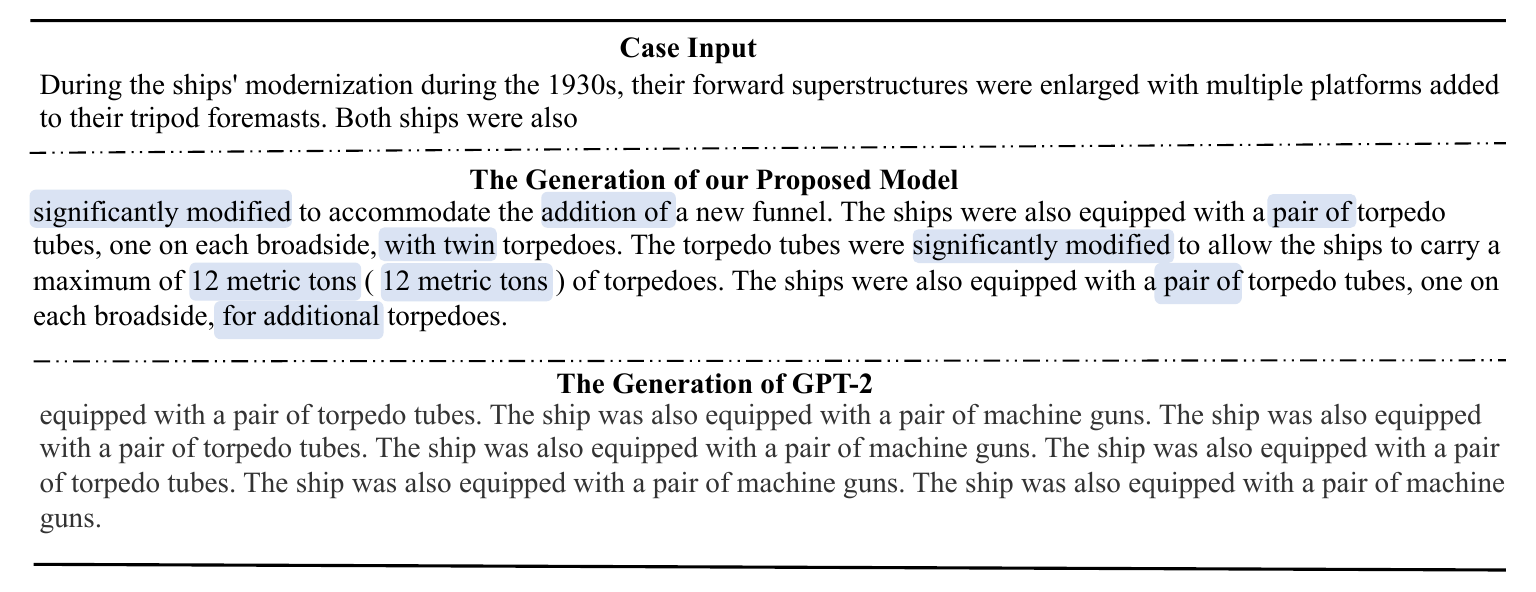}
\caption{A comparison between texts generated by our proposed model and GPT-2. The tokens highlighted in blue are from dynamic vocabulary while others are from fixed token ones.}
\label{case_study}
\end{figure*}
 As illustrated in Figure \ref{case_study}, the continuation of our model consists of both tokens and phrases (such as the phrase ``\emph{significantly modified}'' highlighted in blue at the first decoding step) and its content embodies further details about the modernization of the ship, including the equipment of a pair of torpedo tubes, their positions, and the maximum load. While GPT-2 repeatedly generates completely identical sentences, which is parallel with its low diversity score of 24.30$\%$. More cases are provided in Appendix \ref{appendix: case study}.

\paragraph{Sequence Compression}
 Sequence compression reflects the length of text that a model can accommodate within the same window size. 
 % A higher compression rate can increase the effective text capacity of the model and improve inference efficiency. 
 Following \citet{dagan2024getting}, we measure the two compression metrics, normalized sequence length (NSL) and the average number of Bytes per Token. 
 NSL is the token count of a tokenized sequence from the tokenizer $T$. 
 Given that our model does not incorporate a genuine tokenizer, we take the outputs of each decoding step as the tokenization results. 
 We report scores from tokenizers of GPT-2 and MWT on our model's outputs.
 
\begin{table}[htp]
\begin{center}
\scalebox{0.8}{
\begin{tabular}{ccc}
\toprule
\multicolumn{1}{c}{\bf Model}  & \multicolumn{1}{c}{\bf NLS $\downarrow$} & \multicolumn{1}{c}{\bf UTF-8 Bytes $\uparrow$} \\
\midrule
Transformer & 127.72 & 4.28 \\
MWT & 114.84 & 4.77 \\
 \textbf{Ours} & \bf 101.38 & \bf 5.54 \\
\bottomrule
\end{tabular}}
\end{center}
\caption{Compression on WikiText-103. Since CoG, KNN-LM, and RETRO do not modify the model's tokenizer or input vocabulary, the compression results are the same with the Transformer.}
\label{table:compression on wikitext103}
\end{table}

As shown in the table \ref{table:compression on wikitext103}, our proposed model holds the highest information content per token, averaging 101.38 tokens or phrases per sequence and 5.54 UTF-8 bytes per token, and necessitates fewer tokens or phrases to generate the identical text. 
In other words, 
with an equivalent number of context window sizes, our method encodes a more substantial amount of text. 
This is a natural consequence of the fact that the dynamically added phrases contain more tokens.
% \paragraph{Case Study}

\paragraph{Scale Up}
% For a fair comparison between baselines, we choose GPT-2 as the default backbone. 
For a comprehensive evaluation of our method, we deploy the dynamic vocabulary with TinyLlama \cite{zhang2024tinyllama}, which is a 1.1B LLaMA-style backbone, to assess the performance as the scale of LM increases.
% And we have tried to deploy the dynamic vocabulary with TinyLlama, which is a 1.1B model. 
 As shown in table \ref{table: automatic evaluation on wikitext103 of tinyllama}, our proposed model outperforms Standard TinyLlama with 1.09$\%$ MAUVE and 21.46 $\%$ Diversity, which indicates the better fluency and higher diversity of generation from our method. The results are consistent with the experimental conclusion in section \ref{wikitext103-results} and the preliminary findings indicate the effectiveness of our approach on larger models.

\begin{table}[htp]
\begin{center}
\scalebox{0.7}{
\begin{tabular}{ccccc}
\toprule
\multicolumn{1}{c}{\bf Model}  & \multicolumn{1}{c}{\bf MAUVE $\uparrow$} & \multicolumn{1}{c}{\bf Diversity $\uparrow$} & \multicolumn{1}{c}{\bf Latency(s)$\downarrow$} & \multicolumn{1}{c}{\bf PPL $\downarrow$}\\
\midrule
\bf TinyLlama & 20.64 & 32.53 & 4.92 & 5.20 \\
\bf Ours & \bf 22.54 & \bf 53.99 & \bf 3.82 & 12.88 \\
\bottomrule
\end{tabular}}
\end{center}
\caption{The automatic evaluation on the test set of WikiText-103. In this experiment, we use GPT-2 and TinyLlama to initialize the dynamic phrase encoder and the language model, respectively. We utilize parameter-efficient fine-tuning approach-LoRA on TinyLlama and set r, alpha, and dropout as 8, 32, 0.1, separately. }
\label{table: automatic evaluation on wikitext103 of tinyllama}
\end{table}

% \paragraph{Negative Phrases} \label{negative samples}
\subsection{The Influence of Negative Phrases} \label{negative samples} 
As discussed, we have designed several negative sampling strategies and explored their influence on the generation. As reported in table \ref{table: negative samples and segment algorithm}, we have observed that the choice of the negative phrases method significantly impacts the fluency and quality of the generated text. 

\begin{table}
\centering
\resizebox{1.0\linewidth}{!}{
\begin{tabular}{lccc}
\toprule
 \textbf{Negative Samples} & \textbf{MAUVE} $\uparrow$  & \textbf{Diversity} $\uparrow$ & \textbf{PPL} $\downarrow$\\
\midrule
\bf FMM & & &   \\
$\quad$in-batch & 21.95 & \bf 57.92 & 16.48 \\
$\quad$in-batch + pre-batch & 22.28 & 48.91 & 9.02 \\ 
$\quad$generation & \bf 22.87 & 42.19 &  \bf 6.34 \\ 
$\quad$corpus-retrieval & 21.98  & 41.32 &  6.40 \\ 
$\quad$self-retrieval & 21.65 & 41.67 & 6.39 \\ 
$\quad$self-retrieval + generation & 21.25 & 42.40 & 6.62 \\ 
\midrule
\bf N-words & & &   \\
$\quad$in-batch & 24.67 & \bf 64.15 & 17.01 \\
$\quad$in-batch + pre-batch & 23.98 & 61.80 & 14.60 \\ 
$\quad$generation & 24.99 & 49.03 & 8.51 \\ 
$\quad$self-retrieval  & 24.83 & 48.46 & 8.13 \\ 
$\quad$self-retrieval + generation & \bf 25.69 & 47.44 &  \bf 8.03 \\ 
\midrule
\bf N-ids & & &  \\
$\quad$in-batch & 23.96 & \bf 68.44 & 21.53 \\
$\quad$in-batch + pre-batch & 23.66 & 61.16 & 14.83 \\ 
$\quad$generation & 23.91 & 46.40 &  \bf 8.07 \\ 
$\quad$self-retrieval & 23.64 & 48.38 &  8.36 \\ 
$\quad$self-retrieval + generation & \bf 24.85 & 47.08 & 8.21 \\ 
\bottomrule
\end{tabular}
}
\caption{The automatic evaluation on different negative samples and training samples. During testing, each phrase is constrained to 2-8 tokens. Here, the pre-batch method contains prefixes of gold phrases as well and the number of preceding batches is set to 1.}
\label{table: negative samples and segment algorithm}
\end{table}

% Except for the Pre$\_$batch negative sampling strategy, the other methods exhibit comparable performance. Although the former achieves a lower repetition, its generated outputs exhibit a perplexity approximately three points higher than others. Meanwhile, the Retrieval$\_$samples method achieves similar metric scores but imposes stricter constraints. Specifically, it relies on retrieved examples, which increases algorithmic complexity. Besides, N-words and N-ids got a higher MAUVE than FMM, which indicates that the construction of phrases in training samples has a substantial influence on the MAUVE score.

\begin{itemize}[leftmargin=*]
    \item Specifically, compared with the remaining negative sampling methods, the vanilla in-batch and pre-batch negative sampling methods result in a markedly higher PPL (approximately 10 points and 3 points higher in the FMM setting) 
    \footnote{We have observed that there is a positive correlation between Diversity and PPL, which means that the higher the Diversity, the higher the PPL values tend to be as well. We believe that this phenomenon occurs because the model tends to increase the probability of repeating previous sentences \cite{xu2022learning}, leading to a lower PPL and Diversity.}. 
    The results indicate that strong negative phrases
    are crucial for the model’s generation quality.
    
    \item Regarding generation-based and retrieval-based negative phrases, there is no significant performance difference. However, these methods take additional time costs compared to self-retrieval, as the generation-based approach necessitates continuous generations for the provided phrases, and corpus-retrieval requires retrieving from the related corpus. Self-retrieval method may be optimal in this perspective.

    \item Furthermore, among all negative phrases sampling strategies, the perplexity of the FMM setting is consistently lower than that of the N-words and N-ids ones. This phenomenon occurs perhaps because phrases obtained with FMM are relatively meaningful. Interestingly, the average MAUVE values for the N-words and N-ids are approximately $1\%$ higher than that of FMM. The observation indicates that the way to construct train samples has a substantial influence on the text quality.
\end{itemize}

\subsection{Domain Adaptation}

%Obviously, when deploying our proposed model, the underlying architecture of the language model is kept and one can remove the dynamic phrase encoder or replace the retrieval corpus as needed.

The plug-and-play property of the dynamic phrase encoder
motivates us to explore the performance on 
a different domain in a training-free manner. 
Specifically, we investigate the model trained on the WikiText-103 dataset 
while tested on the LawMT \cite{koehn2017challenges} dataset
which is an English-German translation dataset in the legal domain. 
Following \cite{he2021efficient, alon2022neurosymbolic, lan2023copy}, we treat the English portion of this dataset as a retrieval corpus. 
% To guarantee a fair comparison, we also evaluate the performance of the Transformer model both with and without further fine-tuning on LawMT.
As shown in table \ref{table: automatic evaluation on lawmt},
only equipped with dynamic vocabulary extracted on the target domain, 
the model can 
outperform the transformer fine-tuned on LawMT datasets ($3.29\%$ on MAUVE and $2.78\%$ Diversity).
Thus, the learned phrase encoder could
be an efficient tool for lightweight domain generalization.
We also calculate the sequence compression ratio and conduct GPT-4 Evaluation. The details are in Appendix \ref{GPT2 evaluation}, \ref{appendix: sequence compression on lawmt}.

%It means that our model can generate high-quality text in the specific domain without further fine-tuning.

\begin{table}
\begin{center}
\resizebox{\linewidth}{!}{
\begin{tabular}{ccccc}
\toprule
\multicolumn{1}{c}{\bf Model}  & \multicolumn{1}{c}{\bf MAUVE $\uparrow$}  & \multicolumn{1}{c}{\bf Diversity $\uparrow$} & \multicolumn{1}{c}{\bf Latency(s)$\downarrow$} & \multicolumn{1}{c}{\bf PPL $\downarrow$}\\
\midrule
\bf Transformer w/o FT & 22.97 & 72.12 & 1.03 & \bf 3.21 \\
\bf Transformer w/ FT & 23.06 & 80.21 & \bf 1.02 & 3.54 \\
\bf RETRO & 19.07 & 72.68 & 5.72 & 3.78 \\
\bf KMM-LM$^*$ & 23.32 & 19.85 & - & - \\
\bf CoG & 19.46 & 81.93 & 1.39 & 6.74 \\ 
\bf MWT & 24.55 & 77.45 & 1.10 & 5.38 \\ 
\bf Ours & \bf 26.35 & \bf 82.99 & 1.09 & 7.61 \\ 
\bottomrule
\end{tabular}
}
\end{center}
\caption{The automatic evaluation on Law-MT. In this experiment, we retrieve 512 documents for each sample. To guarantee a fair comparison, we also evaluate the performance of the Transformer model both with and without further fine-tuning on LawMT.}
\label{table: automatic evaluation on lawmt}
\end{table}

\subsection{Generation with Citations}

\begin{table*}[htp]
\begin{center}
\scalebox{0.85}{
\begin{tabular}{lccccc}
\toprule
\multicolumn{1}{l}{\bf Model(shot-1)}  & \multicolumn{1}{c}{\bf Citation$\_$rec} & \multicolumn{1}{c}{\bf Citation$\_$prec}  & \multicolumn{1}{c}{\bf QA-EM} & \multicolumn{1}{c}{\bf QA-F1} & \multicolumn{1}{c}{\bf Rouge-L} \\
\midrule
% \multirow{1}{*}{\bf Vicuna-7B} & 13.36 & 17.56 & 20.11 & 26.48 & 36.22 \\
\multirow{1}{*}{\bf TinyLlama} & 0.62 & 1.54 & 6.00 & 8.78 & 25.43 \\
\midrule
\multirow{1}{*}{\bf ours} & & & & & \\
\quad w/ n-grams & \bf 9.76 & \bf 29.30 & 8.88 & 11.83 & 30.06 \\
\quad w/ parsing & 2.94 & 9.17 & \bf 9.87 & \bf 13.06 & \bf 30.16 \\
\quad w/o phrases & 0.20 & 0.44 & 8.81 & 11.81 & 29.60 \\
\bottomrule
\end{tabular}}
\end{center}
\caption{The automatic evaluation on ASQA. In this experiment, we opt for TinyLlama as the language model to imbue the model with in-context learning capabilities. All baseline models are configured in a one-shot setting, with the number of candidate documents set to 3. Parsing denotes that we use Stanza parser \cite{qi2020stanza} to extract phrases from candidate documents, which ensures that the phrases possess a relatively complete and well-defined meaning.}
\label{table: automatic evaluation on asqa}
\end{table*}

Considering that we can develop a dynamic vocabulary tailored to our needs, and recognizing that each potential phrase is uniquely associated with a specific document, our proposed model is designed to be effectively employed in the generation of citations.
The task is formalized as follows: given a query $q$ and a few documents $D$, the model is required to generate an answer with embedded in-line citations of documents in $D$. We run the experiments on the long-form QA dataset, ASQA \cite{DBLP:conf/emnlp/StelmakhLDC22} further processed by \citet{gao2023enabling}, where candidate documents for each query have already been retrieved.
% ASQA \cite{DBLP:conf/emnlp/StelmakhLDC22} is a long-form QA dataset
%, in which each question is ambiguous and can be answered by Wikipedia. 
% and we utilize the dataset processed by \citet{gao2023enabling}, where candidate documents for each query have already been retrieved. 
We first label each document with a unique ID marker starting from 1 and then extract phrases from documents with the corresponding marker, such as ``\emph{dynamic vocabulary}[1]'' from the document with mark ``[1]''. Therefore, phrases in the generated answers could reflect the citation process.

% during generation, we can easily know which document the generated phrase belongs to, thereby solving the citation task.

\paragraph{Results}
We evaluate the generated results from two perspectives: QA accuracy and citation quality. 
For QA accuracy, we evaluate Exact-Match, F1-score, and Rouge-L and we calculate Recall and Precision in terms of citation quality.  
 Refer to their detailed definitions provided in \citet{gao2023enabling} for an in-depth understanding. Following \cite{gao2023enabling},
we provide the model with the $k$ documents and leverage in-context learning to
instruct it to cite accordingly. 
% We also leverage in-context learning and provide a demonstration in the evaluation process.

The results demonstrate a significant boost in the citation capability of our model with citation recall and precision surpassing TinyLlama baseline by $9.14\%$ and $27.76\%$, respectively. However, phrase collections have a significant impact on the citation results. The phenomenon occurs potentially due to 
the extensive collection of phrases by the n-grams approach and thus more suitable phrases could align with the generated text.
% the property of the n-grams approach, which yields
% an extensive collection of phrases.
% Consequently, there is a higher likelihood of encountering suitable phrases that align with the generated context.

Furthermore, our model exhibits a superior QA performance with an EM score of $9.87\%$ and an F1 of $13.06\%$.
Due to our model's further fine-tuning on WikiText-103 and the property that responding to a query in ASQA necessitates Wikipedia-based information, our model's QA performance is expected to be excellent with the absence of phrases (i.e., the setting of ours w/o phrases).

% when not utilizing phrases (i.e., ours w/o phrases) as compared to directly employing TinyLLama.

%% file: related_work.tex
%auto-ignore
\section{Related Work}

%When building vocabularies, 
%one needs to determine the granularity of tokens. 
%Finer tokens are better for out-of-vocabulary words
%while more ambiguous.
%Coarser tokens are more precise while more affected by OOV.
%Modern tokenizers (e.g., BPE \cite{DBLP:journals/corr/SennrichHB15}) settle the trade-off 
%by redefining a vocabulary size.
%The vocabulary is then frozen during
%construction and deployment of models.

%Despite its fundamental importance, 
%Before any other model designing and training,
%language models should first 
%configure a tokenizer (e.g., BPE \cite{}) to 
%build vocabulary.

%It is predefined before training a language model
%according to a tokenization algorithm and a corpus
%to learn the tokenizer.
%Roughly, modern tokenizers (e.g., BPE \cite{sennrich-etal-2016-neural})
%seek vocabularies that can compress their 
%training corpus as efficiently as possible.

\paragraph{Tokenizer}
Tokenizer is an essential component of language models \cite{dagan2024getting, mielke2021words}, responsible for transforming raw text into a sequence of tokens. 
% Byte-Pair Encoding(BPE) \cite{sennrich2015neural} algorithm is a simple word segmentation method to build tokenizer\cite{radford2019language, liu2019roberta, lewis2019bart, he2021deberta}), that iteratively merges characters into longer units by finding frequently occurring contiguous patterns and there exist other tokenization algorithms, such as Unigram \cite{kudo2018subword} and WordPiece tokenization used in BERT \cite{devlin2019bert}. 
Byte-Pair Encoding (BPE) is commonly used to build tokenizer \cite{radford2019language, liu2019roberta, lewis2019bart, he2021deberta} and, there exist other tokenization algorithms, such as Unigram \cite{kudo2018subword} and WordPiece tokenization used in BERT \cite{devlin2019bert}.
However, these tokenizations are limited to subwords or whole words. \citet{kumar2022bpe} and \citet{Gee_2023} generalize the BPE algorithm to multi-words and multi-tokens separately. Whereas these approaches necessitate training the tokenizer and remain static. 
  % The selection of hyperparameter, vocabulary size, is of significance. According to Zipf's law \cite{zipf2016human}, when vocabulary size is large, the model tends to encounter each token less frequently, potentially leading to negative impacts on downstream performance.
  
  % CoG \cite{lan2023copy}, which employs a "dynamic vocabulary",  retrieves related documents based on the input text and expanded vocabulary with phrases extracted from these documents. 

    CoG \cite{lan2023copy} and CoG-2 \cite{cao2024retrieval} both employ a ``dynamic vocabulary'' by expanding vocabulary with phrases extracted from related documents. 
  However, these two methods only employ dynamic vocabulary in the output module and split phrases into tokens in the input. In this paper, we treated phrases as atomic units same as tokens, and dynamically expanded vocabulary both in input and output layers. 
  
\paragraph{Sequence Compression}
Language models are constrained by the limited length of input sequences they can process. Increasing this length results in a prohibitive computational overhead.
% , growing quadratically with the sequence length \cite{child2019generating, beltagy2020longformer, choromanski2022rethinking}. 
% A promising research direction focuses on reducing the length of input sequences to decrease the computational complexity of large language models. 
A series of techniques have been proposed to compress sentences into one or a few tokens or latent representations
% , thereby increasing the effective input text length for the model 
\cite{10.5555/3618408.3619584, chevalier2023adapting, bulatov2022recurrent, mu2024learning}. 
% CoLLEGe \cite{teehan2024college} employs few-shot examples to obtain embeddings for novel concepts at the phrase level.
MWT \cite{Gee_2023} enhances compression by retraining the tokenizer, incorporating the most frequent n-grams of a support corpus into the vocabulary. 
% Our method also performs phrase-level compression, but unlike CoLLEGe, where each sample contains only one compressed phrase, our samples may contain multiple compressed phrases, further improving the compression rate.
In contrast to the static vocabulary of MWT, our method dynamically adapts the model's vocabulary to the input text, resulting in a more flexible and efficient adaptation.

%% file: appendix.tex
%auto-ignore
\appendix

\section{Full Results} \label{appendix: nucleus results}

We show the full results of our experiments in Tables \ref{table: automatci evaluation on wikitext103 both greedy and nucleus}, \ref{table: negative samples and segment algorithm both greedy and nucleus}, \ref{table: automatic evaluation on lawmt both greedy and nucleus}, \ref{table: negative samples and segment algorithm both greedy and nucleus lawmt}.
\begin{table*}
\begin{center}
\scalebox{0.8}{    
\begin{tabular}{ccccccccc}
\toprule
\multicolumn{1}{c}{\bf Model}  &\multicolumn{1}{c}{\bf Decoding} & \multicolumn{1}{c}{\bf MAUVE $\uparrow$}  & \multicolumn{1}{c}{\bf Rep-2 $\downarrow$} & \multicolumn{1}{c}{\bf Rep-3 $\downarrow$} & \multicolumn{1}{c}{\bf Rep-4 $\downarrow$} & \multicolumn{1}{c}{\bf Diversity $\uparrow$} & \multicolumn{1}{c}{\bf Latency(s)$\downarrow$} & \multicolumn{1}{c}{\bf PPL $\downarrow$}\\
\midrule
\multirow{2}{*}{\bf Transformer} & greedy & 20.47 & 41.96 & 36.82 & 33.74 & 24.30 & 1.10 & 3.60 \\
~ & nucleus &  25.05 & 5.40 & 1.44 & 0.51 & 92.76 & 1.15 & 31.01 \\
\hline
\multirow{2}{*}{\bf RETRO} & greedy & 19.59 & 43.78 & 38.58 & 35.35 & 22.33 & 4.43 & 3.96 \\
~ & nucleus & 20.77 & 5.83 & 1.91 & 0.83 & 91.61 & 5.43 & 39.74 \\
\hline
\multirow{2}{*}{\bf KMM-LM$^*$} & greedy  & 19.92 & 43.79 & 38.76 & 35.69 & 22.13 & 10.36 & 3.48 \\
~ & nucleus & 22.50 & 3.33 & 0.69 & 0.21 & 95.8 & 10.42 & 78.01 \\
\hline
\multirow{2}{*}{\bf CoG} & greedy & 21.61 & 34.77 & 30.67 & 28.35 & 32.41 & 1.04 & 7.89 \\ 
 ~ & nucleus & 25.96 & 5.43 & 1.53 & 0.67 & 92.50 & 1.06 & 36.66 \\
\hline
\multirow{2}{*}{\bf GPT+MWT} & greedy & 24.74 & 33.78 & 26.72 & 22.76 & 37.48 & 1.13 & 5.58 \\ 
 ~ & nucleus & 25.66 & 4.18 & 0.90 & 0.29 & 94.68 & 1.17 & 55.02 \\
\hline
% n-words generate + smaple
% \multirow{2}{*}{\bf Ours} & greedy & 25.69 & 27.77 & 20.80 & 17.08 & 47.44 & 0.99 & 8.03 \\ 
%  ~ & nucleus & 24.34 & 4.59 & 1.03 & 0.28 & 94.16 & 1.00 & 51.38 \\
% n-words sample
\multirow{2}{*}{\bf Ours} & greedy & 25.69 & 27.77 & 20.80 & 17.08 & 47.44 & 0.99 & 8.03 \\ 
 ~ & nucleus & 24.34 & 4.59 & 1.03 & 0.28 & 94.16 & 1.00 & 51.38 \\
\bottomrule
\end{tabular}}
\end{center}
\caption{The automatic evaluation on the test set of WikiText-103. $*$ denotes that the results are obtained from CoG \cite{lan2023copy} paper. For each sample, the first 32 tokens are provided and models are tasked with generating the subsequent 128 tokens. We can observe that our proposed model achieves the best scores in most metrics.}
\label{table: automatci evaluation on wikitext103 both greedy and nucleus}
\end{table*}

\begin{table*}
\begin{center}
\scalebox{0.75}{
\begin{tabular}{lcccccccc}
\toprule
\multicolumn{1}{c}{\bf Negative Samples} & \multicolumn{1}{c}{\bf Decoding}  & \multicolumn{1}{c}{\bf MAUVE $\uparrow$}  & \multicolumn{1}{c}{\bf Rep-2 $\downarrow$} & \multicolumn{1}{c}{\bf Rep-3 $\downarrow$} & \multicolumn{1}{c}{\bf Rep-4 $\downarrow$} & \multicolumn{1}{c}{\bf Diversity $\uparrow$} & \multicolumn{1}{c}{\bf Latency(s)$\downarrow$} & \multicolumn{1}{c}{\bf PPL $\downarrow$}\\
\midrule
\bf FMM & & & & & & &  \\
 \hdashline
 \multirow{2}{*}{$\quad$ in-batch} & greedy & 21.95 & 23.42 & 15.29 & 10.71 & 57.92 & 0.94 & 16.48  \\ 
 ~ & nucleus & 23.17 & 4.17 & 0.92 & 0.29 & 94.67 & 0.84 & 78.20 \\
  \hdashline
  \multirow{2}{*}{$\quad$ pre-batch} & greedy & 22.28 & 26.90 & 20.07 & 16.29 & 48.91 & 0.95 & 9.02 \\ 
 ~ & nucleus & 20.59 & 4.62 & 1.07 & 0.35 & 94.03 & 0.88 & 56.28 \\
  \hdashline
  \multirow{2}{*}{$\quad$generation} & greedy & 22.87 & 31.17 & 23.82 & 19.55 & 42.19 & 1.20 & 6.34 \\ 
 ~ & nucleus & 20.33 & 4.35 & 1.01 & 0.31 & 94.39 & 1.06 & 49.51 \\ 
 \hdashline
 \multirow{2}{*}{$\quad$corpus-retrieval} & greedy & 21.98 & 31.47 & 24.39 & 20.26 & 41.32 & 1.12 & 6.40 \\ 
 ~ & nucleus & 20.52 & 4.36 & 1.00 & 0.32 & 94.38 & 1.08 & 51.60 \\
 \hdashline
 \multirow{2}{*}{$\quad$self-retrieval} & greedy & 21.65 & 31.33 & 24.15 & 20.00 & 41.67 & 1.15 & 6.39 \\ 
 ~ & nucleus & 20.63 & 4.37 & 1.00 & 0.35 & 94.34 & 1.04 & 49.93 \\
 \hdashline
 \multirow{2}{*}{$\quad$self-retrieval + generation} & greedy & 21.25 & 30.89 & 23.73 & 19.57 & 42.40 & 1.16 & 6.62 \\ 
 ~ & nucleus & 20.34 & 4.24 & 0.96 & 0.29 & 94.57 & 1.04 & 52.27 \\
\hline
\bf N-words & & & & & & &  \\
 \hdashline
 \multirow{2}{*}{$\quad$in-batch} & greedy & 24.67 & 20.80 & 12.22 & 7.72 & 64.15 & 0.88 & 17.01 \\ 
 ~ & nucleus & 24.24 & 4.76 & 1.16 & 0.40 & 93.76 & 0.81 & 68.25 \\
  \hdashline
 \multirow{2}{*}{$\quad$pre-batch} & greedy & 23.98 & 19.58 & 13.63 & 11.02 & 61.80 & 1.16 & 14.60 \\ 
 ~ & nucleus & 23.60 & 5.71 & 1.82 & 0.92 & 91.73 & 1.11 & 47.17 \\
  \hdashline
  \multirow{2}{*}{$\quad$generation} & greedy & 24.99 & 26.72 & 19.95 & 16.41 & 49.03 & 0.94 & 8.51 \\ 
 ~ & nucleus & 24.85 & 4.64 & 1.07 & 0.31 & 94.04 & 0.94 & 50.65\\
   \hdashline
 \multirow{2}{*}{$\quad$self-retrieval } & greedy & 24.83 & 27.21 & 20.23 & 16.54 & 48.46 & 0.96 & 8.13 \\ 
 ~ & nucleus & 24.51 & 4.57 & 1.05 & 0.33 & 94.12 & 0.94 & 51.85 \\
   \hdashline
 \multirow{2}{*}{$\quad$self-retrieval + generation} & greedy & 25.69 & 27.77 & 20.80 & 17.08 & 47.44 & 0.99 & 8.03 \\ 
 ~ & nucleus & 24.34 & 4.59 & 1.03 & 0.28 & 94.16 & 1.00 & 51.38 \\
\hline
\bf N-ids & &  & & & & & &  \\
 \hdashline
 \multirow{2}{*}{ $\quad$in-batch} & greedy & 23.96 & 18.63 & 10.30 & 6.22 & 68.44 & 0.81 & 21.53 \\ 
 ~ & nucleus & 23.17 & 4.77 & 1.18 & 0.43 & 93.71 & 0.70 & 81.06 \\
  \hdashline
  \multirow{2}{*}{ $\quad$pre-batch} & greedy & 23.66 & 19.81 & 13.96 & 11.36 & 61.16 & 1.12 & 14.83 \\ 
 ~ & nucleus & 22.84 & 5.17 & 1.52 & 0.67 & 92.77 & 0.92 & 54.52 \\
  \hdashline
  \multirow{2}{*}{$\quad$generation} & greedy & 23.91 & 28.12 & 21.45 & 17.82 & 46.40 & 0.99 & 8.07 \\ 
 ~ & nucleus & 24.50 & 4.41 & 0.97 & 0.29 & 94.38 & 0.96 & 53.98 \\
   \hdashline
 \multirow{2}{*}{$\quad$self-retrieval} & greedy & 23.64 & 27.29 & 20.33 & 16.49 & 48.38 & 1.02 & 8.36 \\ 
 ~ & nucleus & 23.85 & 4.43 & 0.94 & 0.27 & 94.41 & 0.88 & 55.76 \\
   \hdashline
 \multirow{2}{*}{$\quad$self-retrieval + generation} & greedy & 24.85 & 27.85 & 21.04 & 17.36 & 47.08 & 1.01 & 8.21 \\ 
 ~ & nucleus & 23.91 & 4.41 & 0.96 & 0.28 & 94.40 & 0.98 & 53.03 \\
\bottomrule
\end{tabular}}
\end{center}
\caption{The automatic evaluation on different negative samples with greedy and nucleus sampling (top-p: 0.95) decoding algorithms on the WikiText103 dataset. The constructions of training samples and negative phrases have a significant influence on the generated text.}
\label{table: negative samples and segment algorithm both greedy and nucleus}
\end{table*}

% retrieval 512 documents
\begin{table*}
\begin{center}
\scalebox{0.8}{
\begin{tabular}{ccccccccc}
\toprule
\multicolumn{1}{c}{\bf Model}  &\multicolumn{1}{c}{\bf Decoding} & \multicolumn{1}{c}{\bf MAUVE $\uparrow$}  & \multicolumn{1}{c}{\bf Rep-2 $\downarrow$} & \multicolumn{1}{c}{\bf Rep-3 $\downarrow$} & \multicolumn{1}{c}{\bf Rep-4 $\downarrow$} & \multicolumn{1}{c}{\bf Diversity $\uparrow$} & \multicolumn{1}{c}{\bf Latency(s)$\downarrow$} & \multicolumn{1}{c}{\bf PPL $\downarrow$}\\
\midrule
\multirow{2}{*}{\bf Transformer w/o FT} & greedy & 22.97 & 13.36 & 9.69 & 7.84 & 72.12 & 1.03 & 3.21 \\
~ & nucleus & 24.15 & 4.05 & 1.62 & 0.80 & 93.64 & 1.05 & 31.48 \\
\hline
\multirow{2}{*}{\bf Transformer w/ FT} & greedy & 23.06 & 9.74 & 6.45 & 5.00 & 80.21 & 1.02 & 3.54 \\
~ & nucleus & 25.12 & 4.36 & 1.73 & 0.87 & 93.17 & 1.08 & 14.94 \\
\hline
\multirow{2}{*}{\bf RETRO} & greedy & 19.07 & 13.19 & 9.34 & 7.66 & 72.68 & 5.72 & 3.78 \\
~ & nucleus & 21.26 & 3.30 & 1.18 & 0.55 & 95.03 & 5.54 & 57.40 \\
\hline
\multirow{2}{*}{\bf KMM-LM$^*$} & greedy  & 23.32 & - & - & - & 19.85 & - & - \\
~ & nucleus &  24.75 & - & -& - & 94.60 & - & -\\
\hline
\multirow{2}{*}{\bf CoG} & greedy & 19.46 & 9.29 & 5.68 & 4.24 & 81.93 & 1.39 & 6.74 \\ 
 ~ & nucleus & 24.45 & 4.57 & 1.58 & 0.72 & 93.25 & 0.89 & 32.01 \\
\hline
\multirow{2}{*}{\bf GPT+MWT} & greedy & 24.55 & 11.59 & 7.34 & 5.46 & 77.45 & 1.10 & 5.38 \\ 
 ~ & nucleus & 22.68 & 3.15 & 1.01 & 0.39 & 95.49 & 1.16 & 68.55 \\
\hline
% n-words generate-sample
% \multirow{2}{*}{\bf Ours} & greedy & 21.97 & 10.96 & 6.78 & 4.87 & 78.96 & 1.00 & 6.35 \\ 
%  ~ & nucleus & 23.90 & 4.37 & 1.64 & 0.75 & 93.35 & 1.00 & 49.39\\
\multirow{2}{*}{\bf Ours}& greedy & 26.35 & 9.26 & 5.21 & 3.52 & 82.99 & 1.09 & 7.61 \\ 
 ~ & nucleus & 24.80 & 3.63 & 1.17 & 0.48 & 94.78 & 0.93 & 60.70 \\
\bottomrule
\end{tabular}}
\end{center}
\caption{The automatic evaluation on LawMT. We directly retrieve 512 documents for each sample in this experiment. Our proposed model even outperforms the Transformer further fine-tuned on the LawMT corpus.}
\label{table: automatic evaluation on lawmt both greedy and nucleus}
\end{table*}

\begin{table*}
\begin{center}
\scalebox{0.75}{
\begin{tabular}{lcccccccc}
\toprule
\multicolumn{1}{c}{\bf Negative Samples} & \multicolumn{1}{c}{\bf Decoding}  & \multicolumn{1}{c}{\bf MAUVE $\uparrow$}  & \multicolumn{1}{c}{\bf Rep-2 $\downarrow$} & \multicolumn{1}{c}{\bf Rep-3 $\downarrow$} & \multicolumn{1}{c}{\bf Rep-4 $\downarrow$} & \multicolumn{1}{c}{\bf Diversity $\uparrow$} & \multicolumn{1}{c}{\bf Latency(s)$\downarrow$} & \multicolumn{1}{c}{\bf PPL $\downarrow$}\\
\midrule
\bf FMM & & & & & & &  \\
 \hdashline
 \multirow{2}{*}{$\quad$ pre-batch} & greedy & 23.65 & 9.39 & 5.00 & 3.03 & 83.48 & 0.90 & 13.86 \\ 
 ~ & nucleus & 22.73 & 4.82 & 1.87 & 0.85 & 92.60 & 0.84 & 68.31 \\
  \hdashline
  \multirow{2}{*}{$\quad$ pre-batch} & greedy & 25.00 & 8.71 & 4.76 & 3.16 & 84.20 & 0.98 & 8.26 \\ 
 ~ & nucleus & 23.19 & 3.71 & 1.19 & 0.50 & 94.66 & 0.83 & 60.34 \\
  \hdashline
  \multirow{2}{*}{$\quad$generation} & greedy & 22.87 & 11.00 & 6.76 & 4.85 & 78.96 & 1.26 & 6.17 \\ 
 ~ & nucleus & 22.50 & 3.50 & 1.13 & 0.48 & 94.95 & 1.07 & 65.26 \\ 
 \hdashline
 \multirow{2}{*}{$\quad$Retrieval-samples} & greedy & 23.00 & 10.45 & 6.36 & 4.53 & 80.06 & 1.21 & 6.11 \\ 
 ~ & nucleus & 23.24 & 3.43 & 1.01 & 0.46 & 95.07 & 1.02 & 68.26 \\
 \hdashline
 \multirow{2}{*}{$\quad$self-retrieval} & greedy & 23.41 & 10.98 & 6.80 & 4.92 & 78.89 & 1.20 & 6.11 \\ 
 ~ & nucleus & 23.22 & 3.48 & 1.05 & 0.43 & 95.10 & 0.98 & 67.14 \\
 \hdashline
 \multirow{2}{*}{$\quad$self-retrieval + generation} & greedy & 24.15 & 10.50 & 6.31 & 4.49 & 80.08 & 1.22 & 6.24 \\ 
 ~ & nucleus & 22.55 & 3.40 & 1.16 & 0.53 & 94.98 & 1.04 & 69.40 \\
\hline
\bf N-words & & & & & & &  \\
 \hdashline
 \multirow{2}{*}{$\quad$in-batch} & greedy & 24.27 & 10.07 & 5.31 & 3.16 & 82.47 & 0.86 & 15.28 \\ 
 ~ & nucleus & 25.48 & 5.36 & 2.12 & 1.00 & 91.71 & 0.80 & 61.90 \\
  \hdashline
   \multirow{2}{*}{$\quad$pre-batch} & greedy & 26.15 & 6.53 & 3.11 & 1.92 & 88.82 & 0.61 & 14.40 \\ 
 ~ & nucleus & 25.15 & 4.07 & 1.41 & 0.61 & 94.00 & 0.53 & 45.79 \\
  \hdashline
  \multirow{2}{*}{$\quad$generation} & greedy & 26.35 & 9.26 & 5.21 & 3.52 & 82.99 & 1.09 & 7.61 \\ 
 ~ & nucleus & 24.66 & 3.53 & 1.16 & 0.48 & 94.89 & 0.92 & 62.58 \\
   \hdashline
 \multirow{2}{*}{$\quad$self-retrieval } & greedy & 23.65 & 8.92 & 4.88 & 3.29 & 83.87 & 1.04 & 8.05 \\ 
 ~ & nucleus & 24.71 & 3.54 & 1.09 & 0.42 & 95.00 & 0.81 & 62.51 \\
   \hdashline
 \multirow{2}{*}{$\quad$self-retrieval + generation} & greedy & 26.35 & 9.26 & 5.21 & 3.52 & 82.99 & 1.09 & 7.61 \\ 
 ~ & nucleus & 24.80 & 3.63 & 1.17 & 0.48 & 94.78 & 0.93 & 60.70 \\
\hline
\bf N-ids & &  & & & & & &  \\
 \hdashline
 \multirow{2}{*}{ $\quad$in-batch} & greedy & 25.77 & 9.12 & 4.44 & 2.47 & 84.70 & 0.81 & 17.49 \\ 
 ~ & nucleus & 26.04 & 5.19 & 2.06 & 0.95 & 91.98 & 0.70 & 66.18 \\
  \hdashline
  \multirow{2}{*}{ $\quad$pre-batch} & greedy & 25.08 & 6.70 & 3.14 & 1.87 & 88.68 & 0.62 & 14.49 \\ 
 ~ & nucleus & 23.93 & 4.25 & 1.46 & 0.65 & 93.74 & 0.43 & 47.94 \\
  \hdashline
  \multirow{2}{*}{$\quad$generation} & greedy & 22.55 & 9.24 & 5.21 & 3.55 & 82.98 & 1.04 & 8.03 \\ 
 ~ & nucleus & 23.14 & 3.59 & 1.14 & 0.49 & 94.85 & 0.85 & 61.89 \\
   \hdashline
 \multirow{2}{*}{$\quad$self-retrieval} & greedy & 24.63 & 9.46 & 5.43 & 3.71 & 82.44 & 1.05 & 7.86 \\ 
 ~ & nucleus & 24.19 & 3.58 & 1.11 & 0.44 & 94.94 & 0.78 & 63.87 \\
   \hdashline
 \multirow{2}{*}{$\quad$self-retrieval + generation} & greedy & 23.18 & 9.31 & 5.25 & 3.59 & 82.85 & 1.07 & 7.57 \\ 
 ~ & nucleus & 24.63 & 3.57 & 1.10 & 0.46 & 94.93 & 0.87 & 60.32 \\
\bottomrule
\end{tabular}}
\end{center}
\caption{The automatic evaluation on different negative samples with greedy decoding and nucleus sampling(top-p: 0.95) on the LawMT dataset.}
\label{table: negative samples and segment algorithm both greedy and nucleus lawmt}
\end{table*}

\section{More Implementation Details } \label{ref:IMPLEMENTATION DETAILS of baselines}
 The training of our proposed model was carried out on two NVIDIA RTX 3090 GPUs, each with 24GB of memory, over a total of 400,000 training steps. During the training process, we implemented a gradient accumulation step of 2, with a batch size of 4. We also used a linear learning rate schedule with a warmup, alongside the AdamW optimizer \cite{loshchilov2019decoupled}, maintaining the default beta values. The initial learning rate was set at 5e-5. Additionally, we applied gradient clipping with a clipping value of 1.0 to ensure training stability. 
 When conducting nucleus sampling, we set the $p$ to 0.95.

 For each test sample, we retrieve top-k documents that have similar topics with the sample prefix and extract candidate phrases to construct the dynamic vocabulary. In our experiments, the value of k is set to 32 by default and the candidate phrase is restrained to the length of 2-8 tokens.
 
 We initialize the language model with two models of different scales, GPT-2 and Tinyllama \cite{zhang2024tinyllama}, to verify the effectiveness of our proposed method. We employ full-parameter fine-tuning for GPT-2 and LoRA fine-tuning \cite{hu2021lora} for Tinyllama. When fine-tuning TinyLlama with LoRA, we set r as 8 and alpha as 32.

The experiments of MWT in paper \cite{Gee_2023} are conducted on encoder-only models such as BERT \cite{devlin2019bert} and RoBERTa \cite{liu2019roberta}. In our implementation, we modify the foundation model to GPT2 \cite{radford2019language}, a decoder-only model, and add the top 10000 most frequent 2-grams to the original GPT-2 Tokenizer. The embeddings for newly added words are initialized using Fast Vocabulary Transfer (FVT) \cite{DBLP:conf/emnlp/GeeZRT22}. MWT is trained for a total of 150000 steps on the WikiText103 dataset.

\section{More Details of Automatic Evaluation} \label{appendix: automatic evaluation}
In this section, we provide a detailed introduction to the automatic evaluation metrics.
\begin{itemize}
    \item {\bf MAUVE.} \citet{pillutla2021mauve} measures how closely the token distribution in the generated text matches that in human-written text across the entire test set. We follow prior work and leverage the GPT2-large model to generate the scores. In our implementation, the scaling factor is set as 2.0.
    \item {\bf Rep-n.} \citet{welleck2019neural}  measures the repetition at different n-gram levels in the generated text. It is defined as $100 \times (1.0 - \frac{|unique n-gram(x)|}{|total n-gram(x)|})$. Higher Rep-n represents the severe degeneration problem in generations.
    \item {\bf Diversity.} \citet{welleck2019neural} evaluates the variety of generated content, which is formulated as $\prod_{n=2}^4(1- \frac{Rep-n}{100})$. More informative generations get higher Diversity scores.
    \item {\bf Perplexity} is a measure of the uncertainty or difficulty in predicting the next word in a sequence. A lower perplexity score indicates that the model is more certain about its predictions.
\end{itemize}

% \section{Scaling Up} \label{appendix: scale to 1B}
% For a fair comparison between baselines, we choose GPT-2 as the default backbone. And we have tried to deploy the dynamic vocabulary with TinyLlama, which is a 1.1B model. The results in Table \ref{table: automatic evaluation on wikitext103 of tinyllama} are consistent with the experimental conclusion in the section \ref{wikitext103-results} above.

% \begin{table*}[htp]
% \begin{center}
% \scalebox{0.85}{
% \begin{tabular}{cccccccc}
% \hline
% \multicolumn{1}{c}{\bf Model}  & \multicolumn{1}{c}{\bf MAUVE $\uparrow$}  & \multicolumn{1}{c}{\bf Rep-2 $\downarrow$} & \multicolumn{1}{c}{\bf Rep-3 $\downarrow$} & \multicolumn{1}{c}{\bf Rep-4 $\downarrow$} & \multicolumn{1}{c}{\bf Diversity $\uparrow$} & \multicolumn{1}{c}{\bf Latency(s)$\downarrow$} & \multicolumn{1}{c}{\bf PPL $\downarrow$}\\
% \hline
% \bf TinyLlama & 20.64 & 35.68 & 30.36 & 27.39 & 32.53 & 4.92 & 5.20 \\
% \bf Ours & \bf 22.54 & \bf 24.63 & \bf 17.31 & \bf 13.27 & \bf 53.99 & \bf 3.82 & 12.88 \\
% \hline
% \end{tabular}}
% \end{center}
% \caption{The automatic evaluation on the test set of WikiText-103.}
% \label{table: automatic evaluation on wikitext103 of tinyllama}
% \end{table*}

\section{Real-time Adaptability} \label{appendix: Real-time adaptability}
We have attempted to verify the efficiency when the proposed model adapts its vocabulary in real-time scenarios where new phrases continuously emerge. We give a simulated experiment with dynamic vocabulary updates in real time. Specifically, we first use a document retriever to retrieve top-k-related documents for each given prefix. Then, the candidate phrases P are collected from these documents for selection. Unlike the full off-line computation (the setting in section \ref{wikitext103-results}), we gradually expand the vocabulary during the model's generation. Specifically, we added $5\%$ of the phrases from P to the vocabulary for every 10 tokens generated. 

\begin{table*}[htp]
\begin{center}
\scalebox{0.85}{
\begin{tabular}{cccccccc}
\toprule
\multicolumn{1}{c}{\bf Settings}  & \multicolumn{1}{c}{\bf MAUVE $\uparrow$}  &  \multicolumn{1}{c}{\bf Diversity $\uparrow$} & \multicolumn{1}{c}{\bf Latency(s)$\downarrow$} & \multicolumn{1}{c}{\bf PPL $\downarrow$}\\
\midrule
\bf Ours(70) & 25.27 & 46.11 & 1.03 & 7.78 \\
\bf Ours(70) + real-time & 24.42 & 47.05 & 1.31 & 7.99 \\
\bf Ours(100) & 25.69 & 47.44 & 0.99 & 8.04 \\
\bottomrule
\end{tabular}}
\end{center}
\caption{The results of real-time adaptability. (x) represents that we construct dynamic vocabulary with $x\%$ of P and real-time denotes the real-time scenarios.}
\label{table: real-time adaptability}
\end{table*}

Obviously, the computational and memory costs are linear to the size of on-demand vocabularies, which we believe is reasonable since 1) the encoding of phrases could be computed in the way of parallel and off-line; 2) the prediction over the new phrase table could also be paralleled using the tilling trick \cite{milakov2018online}; 3) in practice, the size of dynamic vocabulary could be controlled by dynamically off-loading unused phrases. As shown in table \ref{table: real-time adaptability}, the increase in latency can be successfully controlled.

\section{Memory and computational resources} \label{appendix: Memory and computational resources}

We control the number of phrases in dynamic vocabulary to illustrate its impact on total FLOPs required to generate text of the same number of tokens after being tokenized by GPT-2.

Despite the addition of 65,536 phrases (more than 50,257 tokens in GPT-2), our model can still save a significant amount of FLOPs compared to the baseline (phrase number = 0 in this table).

\begin{table}[htp]
\begin{center}
\scalebox{0.65}{
\begin{tabular}{lccc}
\toprule
 \multicolumn{1}{c}{\bf Phrase num} & \multicolumn{1}{c}{\bf FLOPS (Rel) (T)} & \multicolumn{1}{c}{\bf Avg Tokens}  & \multicolumn{1}{c}{\bf Memory (Rel) (GB)} \\
\midrule
0 & 4.07(1$\times$) & 128 & 1.2411(1.00$\times$)\\
32 & 2.63(0.65 $\times$) & 88 & 1.2412(1.00 $\times$) \\
128 & 2.06(0.51 $\times$) & 98 & 1.2415(1.00 $\times$) \\
2048 & 2.12(0.52 $\times$) & 95 & 1.2529(1.01 $\times$) \\
8192 & 1.98(0.49 $\times$) & 96 & 1.2880(1.04 $\times$) \\
16384 & 2.39(0.59 $\times$) & 89 & 1.3349(1.08 $\times$) \\
65536 & 2.64(0.65 $\times$) & 73 & 1.6161(1.30 $\times$) \\
\bottomrule
\end{tabular}}
\end{center}
\caption{ The impacts of dynamic Vocabulary on FLOPs and Memory occupation. }
\label{table: Memory and computational resources}
\end{table}

The following is a theoretical analysis.
\paragraph{Memory Overhead.}
The additional memory overhead mainly involves the memory occupation of the dynamic phrase encoder and the phrase embedding. The former is fixed and the latter is linearly related to the number of new phrases added.
Assuming that the memory occupation of phrase encoder and language model is $M_p$ and $M_l$ separately, then the proportion of additional memory overhead is as follows: $M_p + p*d*4B /(M_p+p*d*4B + M_l)$. p is the number of newly added phrases and d denotes the dimension of token embeddings. Therefore, different sizes of language models lead to varying overheads and the overhead is trivial when choosing a larger model, such as Tinyllama.

\paragraph{Computational cost.}Compared to the Transformer, our proposed model requires additional computation on output embeddings during one-step generation: $2pdn$ (n represents the sentence length). Since phrase embeddings can be obtained offline, this item is excluded from the computational cost.

The computational cost of a single forward propagation is $2(n(V+p)d+(24nd^2+4n^d)L)$. And V is the vocabulary size of the language model and L notes the layer numbers.

Therefore, the percentage of additional computational resources for one forward propagation is $p/(V+p+(12d+2n)L)$.

When the dynamic phrase encoder is set as GPT2(124M) and the Language model is initialized with Tinyllama(1.1B), then the percentage of additional memory and computational resources is approximately 10% and 5% (when phrase numbers are 32000, same with the vocabulary size with Tinyllama). Note: To simplify the calculation, we have not taken into account the intermediate variables, KV cache, biases, and other factors. This table mainly focuses on the inference process. When it comes to training, the additional memory/computational resources required for the dynamic vocabulary are nearly subtle when Tinyllama serves as LM.

Although our model will increase minor computational costs on one-step generation, more than one forward process can be saved when generating a phrase with two or more tokens.

\section{Case Study} \label{appendix: case study}

\begin{figure*}[htbp]
\centering
\includegraphics[scale=0.57]{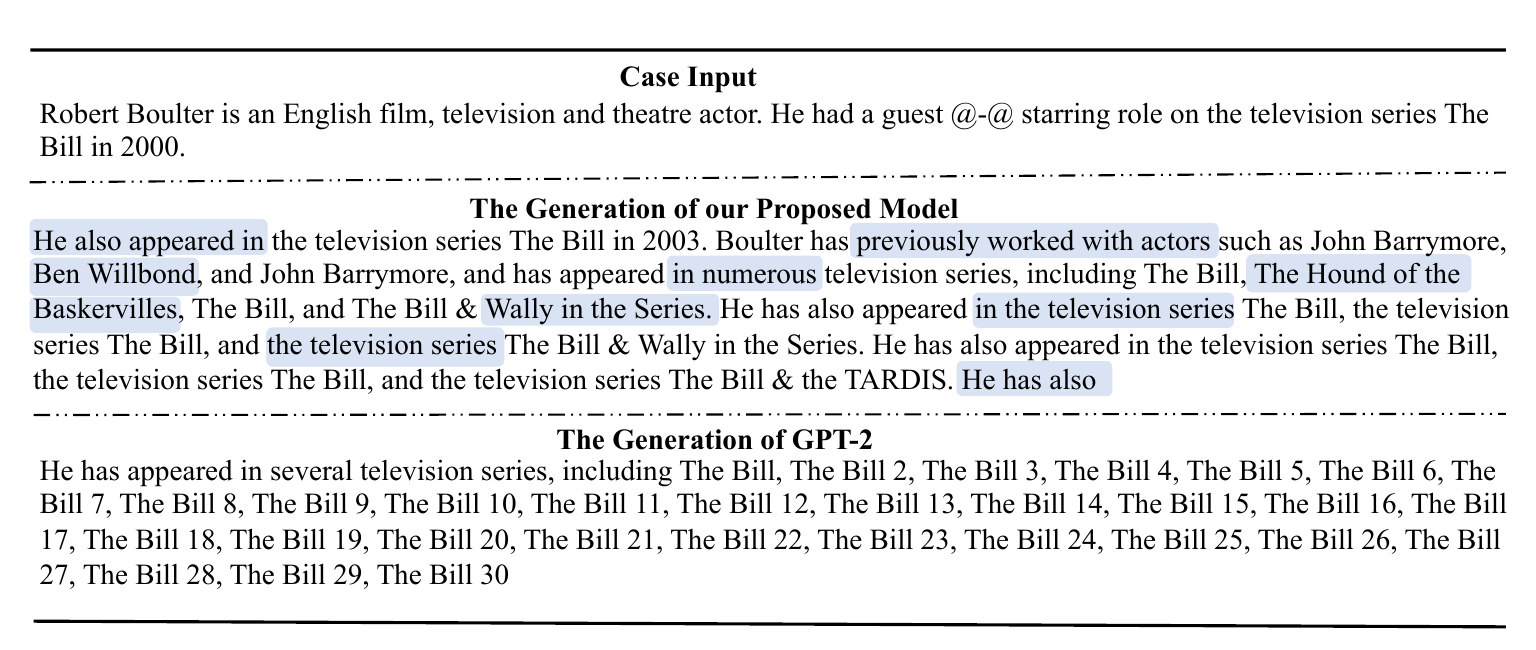}
\caption{A comparation between texts generated by our proposed model and GPT-2. The tokens highlighted in blue are from dynamic vocabulary while others are from fixed token ones.}
\label{case_study_2}
\end{figure*}

\begin{figure*}[htbp]
\centering
\includegraphics[scale=0.57]{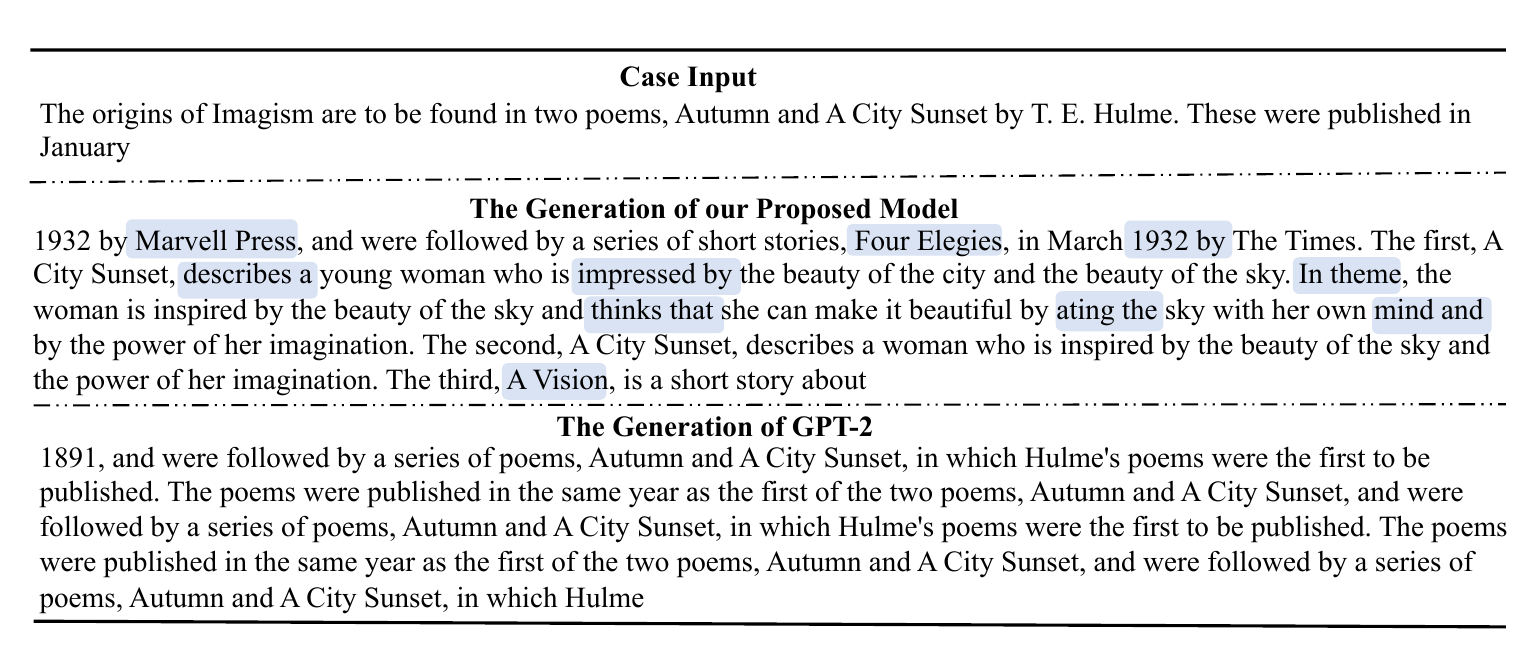}
\caption{A comparation between texts generated by our proposed model and GPT-2. The tokens highlighted in blue are from dynamic vocabulary while others are from fixed token ones.}
\label{case_study_3}
\end{figure*}

In this section, we present some generated examples of our proposed model and GPT-2. As illustrated in Figure \ref{case_study_2} and \ref{case_study_3}, it can be observed that the generations of our model are more informative and more diverse than those of GPT-2.  For example, as shown in Figure \ref{case_study_2}, our content introduces the television series played by Boulter and the actors co-played with Boulter while GPT-2 merely repeats the TV series ``\emph{The Bill}''. Moreover, Figure \ref{case_study_3} presents that the generated text from our proposed model describes richer features about each series than GPT-2.

\section{GPT-4 Evaluation} \label{GPT2 evaluation}

\begin{figure*}[hbpt]
  \centering
  \includegraphics[scale=0.43]{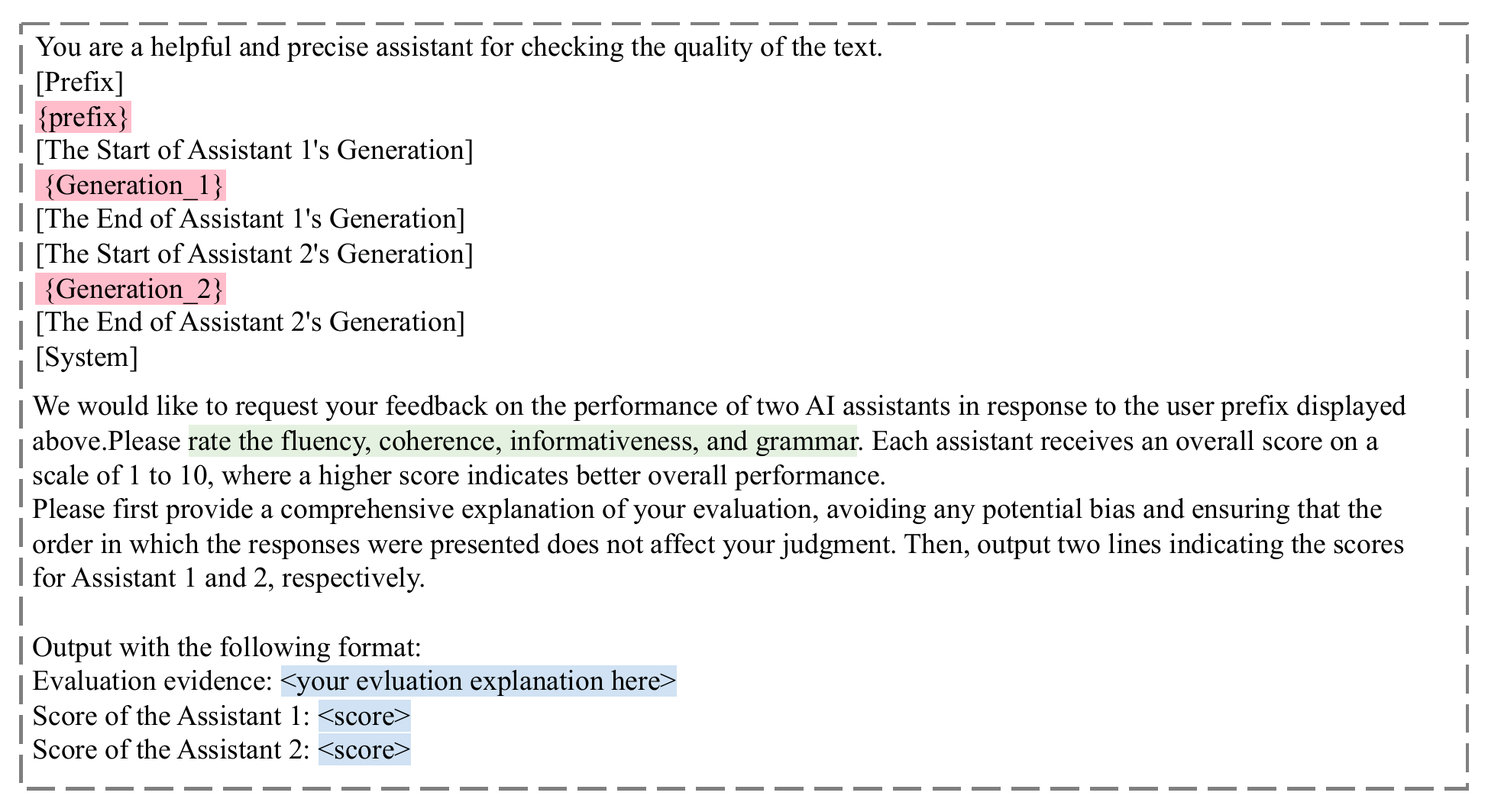}
  \caption{The GPT-4 evaluation template with three slot \{prefix\}, \{Generation$\_$1\} and \{Generation$\_$2\}.}
  \label{fig:gpt_prompt}
\end{figure*}

    Although human evaluation is considered the gold standard for assessing human preferences, it is slow and costly. \citet{zheng2023judging} have demonstrated that strong LLMs, such as GPT-4, can match most human preferences well
, achieving over $80\%$ agreement, which is the same level of agreement between humans. 
Therefore, LLM-as-a-judge is an interpretable approach to approximating human preferences.
 We random sample 100 cases and evaluate the results of the Baselines and our model. GPT-4 is asked to evaluate the generated texts by considering fluency, coherence, informativeness, and grammar. Owing to GPT4's sensitivity to the order of the two candidate sentences \cite{wang2023large}, we adhere to the approach employed in \citet{wang2023large} and determine the final result by calculating the average of the outcomes from interchanging the order of the candidate sentences.

Figure \ref{fig:gpt_prompt} shows the detailed prompt used for GPT-4. Despite the template emphasizing that the order should not affect the results (red text), large language models still exhibit a significant positional bias. Therefore, for each triplet (\emph{prefix, <generation$\_$1>, <generation$\_2$>}), we include another corresponding triplet (\emph{prefix, <generation$\_2$>, <generation$\_1$>}). This is done to mitigate the impact of the order of the two generations on GPT-4 evaluation.
% \label{gpt evaluation}

Table \ref{table: all gpt evaluation} is the full results of our evaluation using GPT-4. It can be seen that our model is capable of producing generations that are comparable or even superior to the baselines.

\begin{table}[htp]
\begin{center}
\scalebox{0.8}{
\begin{tabular}{lccc}
\toprule
 \multicolumn{1}{c}{\bf Comparison (VS)} & \multicolumn{1}{c}{\bf Better} & \multicolumn{1}{c}{\bf No Prefer}  & \multicolumn{1}{c}{\bf Worse} \\
\midrule
\bf WikiText103 & & & \\
\quad  Transformer & 0.61 & 0.05 & 0.34\\
\quad  MWT & 0.58 & 0.02 & 0.40 \\
\quad  CoG & 0.58 & 0.08 & 0.34 \\
\hline
\bf LawMT & & & \\
\quad  Transformer & 0.46 & 0.02 & 0.52 \\
\quad  MWT & 0.67 & 0.07 & 0.26 \\
\quad  CoG & 0.50 & 0.05 & 0.45
\\
\bottomrule
\end{tabular}}
\end{center}
\caption{GPT-4 evaluation on WikiText-103. Due to the sensitivity of GPT-4 to the order of two candidates, we got the final result by calculating the average scores by changing the order of the two candidates.}
\label{table: all gpt evaluation}
\end{table}

\section{Sequence Compression On LawMT} \label{appendix: sequence compression on lawmt}
\begin{table}[ht]
\begin{center}
\scalebox{0.8}{
\begin{tabular}{lcc}
\toprule
\multicolumn{1}{c}{\bf Model}  & \multicolumn{1}{c}{\bf NLS } & \multicolumn{1}{c}{\bf UTF-8 Bytes} \\
\midrule
\multirow{1}{*}{\bf WikiText103} &  & \\
\quad Transformer & 127.72 & 4.28 \\
\quad MWT & 114.84 & 4.77 \\
 \quad Ours & 101.38 & 5.54 \\
\midrule
\multirow{1}{*}{\bf LawMT} &  & \\
\quad Transformer & 128.79 & 5.22 \\
\quad MWT & 124.94 & 5.39 \\
 \quad Ours & 105.38 & 6.53 \\
\bottomrule
\end{tabular}}
\end{center}
\caption{Compression on WikiText-103 and LawMT. Our model compresses text in a larger margin than MWT in the specific domain.}
\label{table:compression on lawmt}
\end{table}
Analogous to the section \ref{wikitext103-results}, we calculate the compression ratio of LawMT. The conclusion aligns with those from section \ref{wikitext103-results}, indicating that our model could yield the highest information density per token. And for an equal number of tokens, our model encompasses a longer effective text length.